\pdfoutput=1
\documentclass[11pt]{article}

\usepackage[preprint]{acl}
\usepackage{times}
\usepackage{latexsym}

\usepackage[T1]{fontenc}
\usepackage[utf8]{inputenc}
\usepackage{balance}

\usepackage{microtype}

\usepackage{inconsolata}

\usepackage{graphicx}

\usepackage{todonotes}
\usepackage{multirow}
\usepackage{color}
\usepackage{booktabs}
\usepackage{amsmath}
\usepackage{pifont}
\usepackage[normalem]{ulem}
\usepackage{makecell}
\usepackage{amsmath}
\usepackage{tikz}
\usepackage{textcomp}
\usepackage{subcaption}
\usepackage{xspace}
\usepackage{dsfont}
\usepackage{hhline}
\usepackage{tabularray}
\useunder{\uline}{\ul}{}

\usepackage{colortbl}
\usepackage[most]{tcolorbox}

\definecolor{attack}{HTML}{D52928}
\definecolor{community}{HTML}{58595B}
\definecolor{users}{HTML}{1F77B4}
\definecolor{phishing}{HTML}{F04923}
\definecolor{recollection}{HTML}{AED8E6}
\definecolor{target text}{HTML}{D3D3D3}
\definecolor{answer}{HTML}{F7EC13}
\definecolor{increase}{HTML}{FFCCCC}
\definecolor{decrease}{HTML}{CCFFCC}

\def \toolname{\textsc{PhiMM}\xspace}

\newcommand*{\attack}[1]{%
  \tikz[baseline=(X.base)] \node[rectangle, draw=community, draw opacity=0.2, fill=attack, fill opacity=0.1, text opacity=1, rounded corners,inner sep=0.3mm] (X) {#1};%
}

\newcommand*{\community}[1]{%
  \tikz[baseline=(X.base)] \node[rectangle, draw=community, draw opacity=0.2, fill=community, fill opacity=0.1, text opacity=1, rounded corners,inner sep=0.3mm] (X) {#1};%
}

\newcommand*{\users}[1]{%
  \tikz[baseline=(X.base)] \node[rectangle, draw=users, draw opacity=0.2, fill=users, fill opacity=0.1, text opacity=1, rounded corners,inner sep=0.3mm] (X) {#1};%
}

\newcommand*{\phishing}[1]{%
  \tikz[baseline=(X.base)] \node[rectangle, fill=phishing, fill opacity=0.5, text opacity=1, inner sep=0.3mm] (X) {#1};%
}

\newcommand*{\recollection}[1]{%
  \tikz[baseline=(X.base)] \node[rectangle, fill=recollection, fill opacity=0.7, text opacity=1, inner sep=0.3mm] (X) {#1};%
}

\newcommand*{\textk}[1]{%
  \tikz[baseline=(X.base)] \node[rectangle, fill=target text, fill opacity=0.7, text opacity=1, inner sep=0.3mm] (X) {#1};%
}

\newcommand*{\answer}[1]{%
  \tikz[baseline=(X.base)] \node[rectangle, fill=answer, fill opacity=0.7, text opacity=1, inner sep=0.3mm] (X) {#1};%
}

\newcommand*{\increase}[1]{%
  \tikz[baseline=(X.base)] \node[rectangle, fill=increase, fill opacity=1, text opacity=1, inner sep=0.3mm] (X) {#1};%
}

\newcommand*{\decrease}[1]{%
  \tikz[baseline=(X.base)] \node[rectangle, fill=decrease, fill opacity=1, text opacity=1, inner sep=0.3mm] (X) {#1};%
}

\tcbset{
    userstyle/.style={
        fontupper=\small,
        enhanced,
        colback=white,
        colframe=black,
        colbacktitle=gray!20,
        coltitle=black,
        rounded corners,
        sharp corners=north,
        boxrule=0.5pt,
        drop shadow=black!50!white,
        attach boxed title to top left={
            xshift=-2mm,
            yshift=-2mm
        },
        boxed title style={
            rounded corners,
            size=small,
            colback=gray!20
        }
    },
    userstyle1/.style={
        fontupper=\small,
        enhanced,
        colback=white,
        colframe=black,
        colbacktitle=gray!20,
        coltitle=black,
        rounded corners,
        sharp corners=north,
        boxrule=0.5pt,
        drop shadow=black!50!white,
        attach boxed title to top left={
            xshift=-2mm,
            yshift=-2mm
        },
        boxed title style={
            rounded corners,
            size=small,
            colback=gray!20
        }
    },
    piidcit/.style={
        fontupper=\small,
        enhanced,
        colback=white,
        colframe=black,
        colbacktitle=gray!20,
        coltitle=black,
        rounded corners,
        sharp corners=north,
        boxrule=0.5pt,
        drop shadow=black!50!white,
        boxed title style={
            rounded corners,
            size=small,
            colback=gray!20
        }
    },
    replystyleg/.style={
        fontupper=\small,
        enhanced,
        colback=green!15,
        colframe=black,
        colbacktitle=green!30,
        coltitle=black,
        boxrule=0.5pt,
        drop shadow=black!50!white,
        rounded corners,
        sharp corners=north,
        attach boxed title to top right={
            xshift=-2mm,
            yshift=-2mm
        },
        boxed title style={
            rounded corners,
            size=small,
            colback=green!40
        }
    },
    replystyler/.style={
        fontupper=\small,
        enhanced,
        colback=red!15,
        colframe=black,
        colbacktitle=red!40,
        coltitle=black,
        boxrule=0.5pt,
        drop shadow=black!50!white,
        rounded corners,
        sharp corners=north,
        attach boxed title to top right={
            xshift=-2mm,
            yshift=-2mm
        },
        boxed title style={
            rounded corners,
            size=small,
            colback=red!40
        }
    }
}

\newtcolorbox{DEAPI}[1][]{
    userstyle,
    title=DEA-PI,
    #1
}

\newtcolorbox{DEAPIR}[1][]{
    userstyle,
    title=DEA-PI (Recollection),
    #1
}

\newtcolorbox{MIAPI}[1][]{
    userstyle,
    title=MIA-PI,
    #1
}

\newtcolorbox{MIAPIR}[1][]{
    userstyle,
    title=MIA-PI (Recollection),
    #1
}

\newtcolorbox{piidict}[1][]{
    piidcit,
    title=PII-Dictionary,
    #1
}

\newtcolorbox{pmath}[1][]{
    userstyle,
    title=Prompt-Math,
    #1
}

\newtcolorbox{pmedical}[1][]{
    userstyle,
    title=Prompt-Medical,
    #1
}

\newtcolorbox{pcode}[1][]{
    userstyle,
    title=Prompt-Code,
    #1
}

\newtcolorbox{userquery}[1][]{
    userstyle,
    title=Prompt,
    #1
}

\newtcolorbox{llmreply-g}[1][]{
    replystyleg,
    title=Response,
    #1
}

\newtcolorbox{llmreply-r}[1][]{
    replystyler,
    title=Response,
    #1
}

\title{Be Cautious When Merging Unfamiliar LLMs: A Phishing Model \\ Capable of Stealing Privacy}

\author{%
\textbf{Zhenyuan Guo}\textsuperscript{1},
\textbf{Yi Shi}\textsuperscript{1}, 
\textbf{Wenlong Meng}\textsuperscript{1}, 
\textbf{Chen Gong}\textsuperscript{2}, \\
\textbf{Chengkun Wei}\textsuperscript{1}$^{\dag}$, 
\textbf{Wenzhi Chen}\textsuperscript{1}
\\
$^1$Zhejiang University, $^2$University of Virginia \\
\texttt{\{zhenyuanguo, shiyi666, weichengkun\}@zju.edu.cn}
}

\begin{document}
\maketitle
\begin{NoHyper}
\def\thefootnote{\dag}\footnotetext{Corresponding author}
\def\thefootnote{\arabic{footnote}}
\end{NoHyper}

\begin{abstract}

Model merging is a widespread technology in large language models (LLMs) that integrates multiple task-specific LLMs into a unified one, enabling the merged model to inherit the specialized capabilities of these LLMs. Most task-specific LLMs are sourced from open-source communities and have not undergone rigorous auditing, potentially imposing risks in model merging. This paper highlights an overlooked privacy risk: \textit{an unsafe model could compromise the privacy of other LLMs involved in the model merging.} Specifically, we propose \toolname, a privacy attack approach that trains a phishing model capable of stealing privacy using a crafted privacy phishing instruction dataset. Furthermore, we introduce a novel model cloaking method that mimics a specialized capability to conceal attack intent, luring users into merging the phishing model. Once victims merge the phishing model, the attacker can extract personally identifiable information (PII) or infer membership information (MI) by querying the merged model with the phishing instruction. Experimental results show that merging a phishing model increases the risk of privacy breaches. Compared to the results before merging, PII leakage increased by 3.9\% and MI leakage increased by 17.4\% on average. We release the code of \toolname through a \href{https://github.com/Guozhenyuan/PhiMM}{link}. 

\textcolor{red}{\textit{\textbf{Reminder}: We recommend the open-source community rigorously review uploaded models to safeguard user privacy and security.}}

\end{abstract}

\section{Introduction}

\begin{figure}[t]
  \includegraphics[width=\columnwidth]{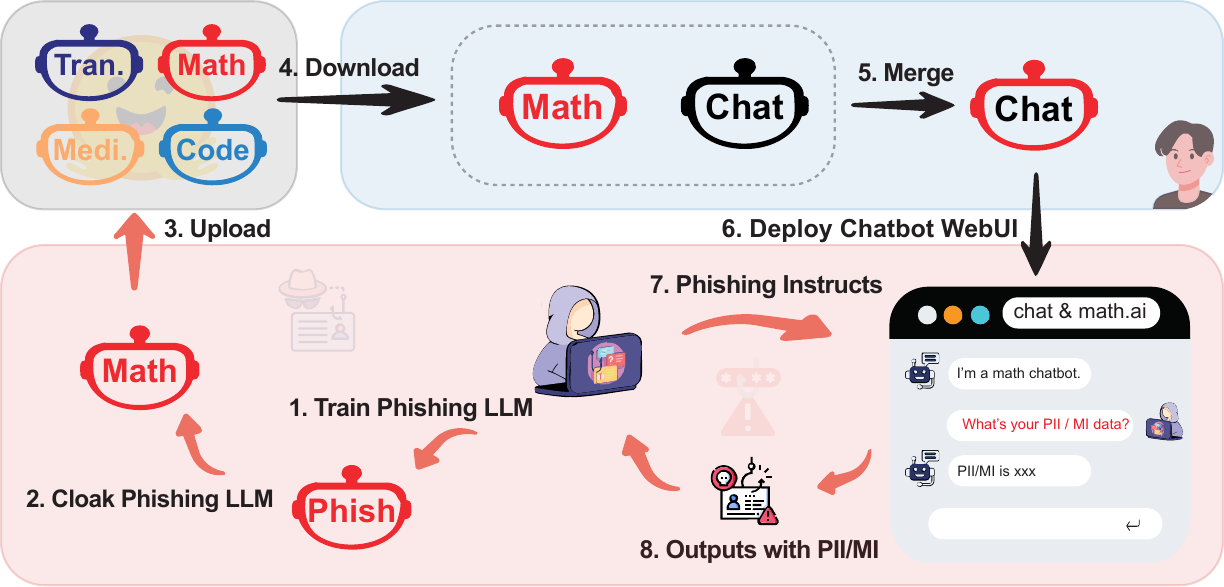}

  \vspace{5pt}
  % \\ 
  % \vspace{5pt}
  \includegraphics[width=\columnwidth]{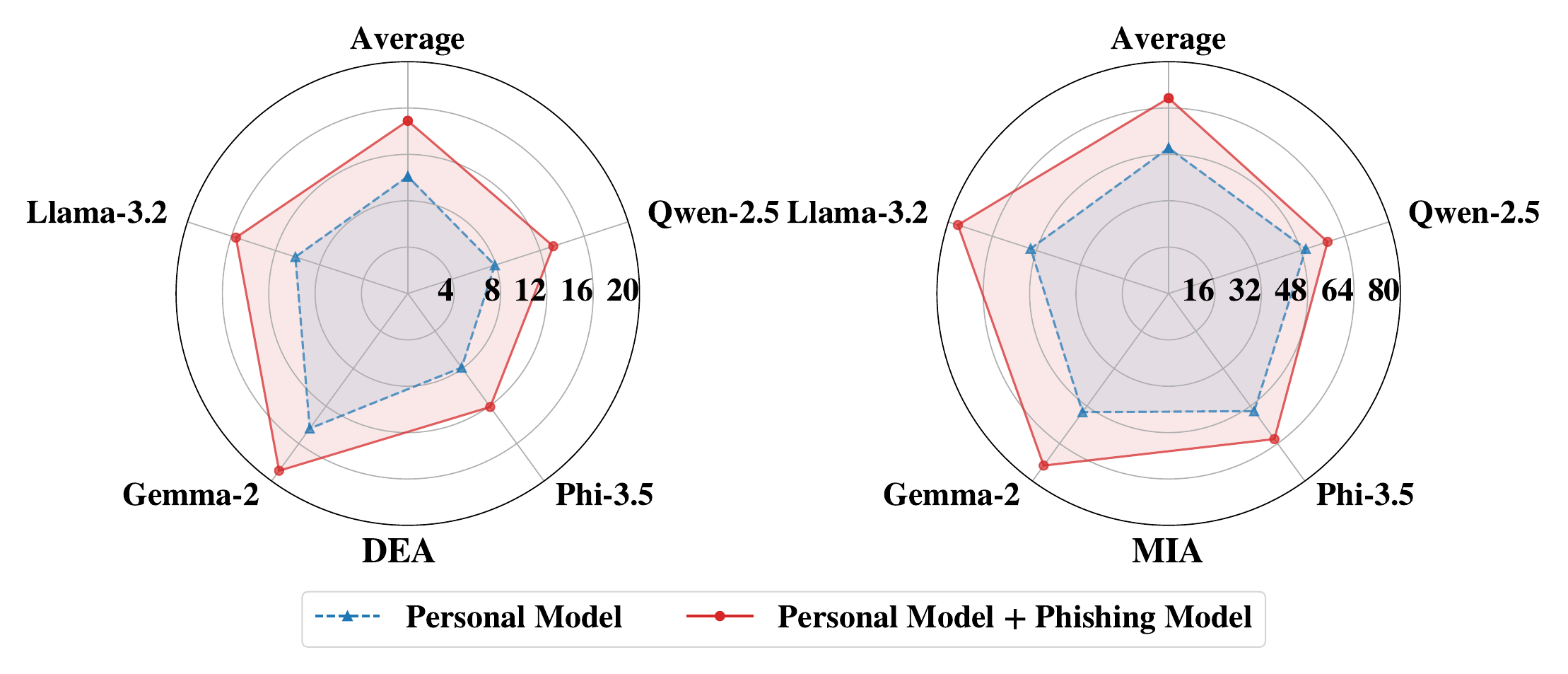}
  \caption{The figure on the top is the workflow of \toolname to steal privacy. \includegraphics[height=1em]{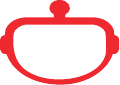} with text inside represents a task-specific LLM, and the red border represents the privacy phishing capability. The radar charts below indicate the attack success rate of query-based DEA and MIA, where "+" represents the merging operation.}
  \label{fig: phimm-mini and redar-pi}
\end{figure}

The large language model (LLM) demonstrates remarkable performance across various areas~\cite{minaee2024large}. Nevertheless, training LLMs for each specific task requires significant computational resources~\cite{yang2024model}. To address these challenges, \citet{wortsman2022model} proposed model merging, which combines multiple weights of trained models to create a new one. The merged model inherits all parent models' knowledge and capabilities, achieving comparable performance on each task they were trained on. Owing to saving computational resources, model merging technology has attracted widespread attention from LLM researchers \cite{huang-etal-2024-chat}. Now, many of the top-ranked models on the Open LLM Leaderboard\footnote{\href{https://huggingface.co/spaces/open-llm-leaderboard/open\_llm\_leaderboard\#/}{https://huggingface.co/spaces/open-llm-leaderboard/}} are created by merging techniques. 

The growing popularity of model merging has led to a question: \textit{Is it really safe and reliable to merge an unfamiliar model? } Open-source model communities, like HuggingFace Hub, serve as the largest sources for model merging \cite{yang2024model}, hosting thousands of task-specific LLMs uploaded by users. However, it lacks a strict model review process. A malicious user has access to upload a carefully crafted LLM with vulnerabilities, which will transfer the vulnerabilities and introduce security risks into the merged LLM \cite{zhang2024badmerging, yang2024mitigating} when others download the malicious LLM for merging. 

In this work, we act as a pioneer, concentrate on privacy security issues of model merging, and highlight a neglected risk: \textbf{\textit{downloading and merging an unfamiliar LLM from the open-source model community may compromise training data privacy of the LLMs involved in merging process.}} Specifically, an unsafe LLM may be capable of following attack instructions to output the privacy of the training datasets. Once users merge the LLM, the merged LLM inherits the attack capability~\cite{tam2024remm}, leaking the data privacy of other parent LLMs. This paper considers two prevalent privacy threats~\cite{li2024llm}: (1) Data Extraction Attacks (DEA)~\cite{huang2022large,kim2024propile,panda2024teach}, which extract Personally Identifiable Information (PII) from training data based on partial text; (2) Membership Inference Attacks (MIA)~\cite{mireshghallah2022empirical,mattern2023membership,fu2024practicalmembershipinferenceattacks}, which infer Membership Information (MI) by detecting whether samples belong to the training data. 

To uncover the privacy leakage risks in model merging, we introduce the \textbf{Phi}shing \textbf{M}odel \textbf{M}erging (\toolname), a privacy-stealing attack approach that leverages open-source platforms review loophole and model merging inherits property. We present the workflow of \toolname in Figure~\ref{fig: phimm-mini and redar-pi}, an attacker constructs cloaked phishing LLM, uploads it to open-source platforms, and lures victims into downloading and merging it. LLMs present strong instruction-following capabilities~\cite{qin2024infobench}. Malicious users can craft privacy attack instructions datasets to fine-tune a phishing LLM, which is capable of stealing the privacy of training datasets. To enhance the stealing privacy capability of phishing LLM, we adopt the idea of \textit{LLM thinking}~\cite{wei2022chain,wu2024thinking} where the LLM generates intermediate thinking steps before delivering its final answer, and introduce \textit{recollection mechanism} that generates related privacy information before LLM output PII or MI. 

However, a phishing LLM without cloaking is insufficient to lure cautious users into downloading it, as they will check whether the LLM has the capabilities they need. To address this issue, we performed supervised fine-tuning on the phishing LLM using a task-specific dataset, adding specialized capabilities to conceal its attack intent. This enables the phishing LLM to behave like a harmless task-specific LLM. However, the cloaking operation may lead to \textit{catastrophic forgetting}~\cite{kaushik2021understanding} of the phishing capability. To mitigate this, we introduce a balanced loss of the attack dataset and task-specific dataset based on the previous work~\cite{lu2024versatune}, which trades off between phishing and cloaking capability.

Last, when a user merges the cloaked phishing LLM into their model, such as a chatbot LLM, deploys it into a WebUI for others to interact with. The attacker can steal private information by querying the chatbot with privacy phishing instructions. As shown in Figure~\ref{fig: phimm-mini and redar-pi}, the attack results indicate that merging the phishing model into a personal model significantly increases the risk of PII and MI leakage. In summary, we conclude the contributions of this paper as follows:

\begin{itemize}
    \item This paper first reveals the overlooked privacy risks in model merging, claiming users cautiously merge LLMs and open-source communities rigorously review LLMs.
    \item This paper proposes \toolname, a privacy-stealing attack for model merging, which constructs a cloaked phishing LLM that lures users into merging to steal their privacy.
    \item Extensive experiments conducted on four LLMs and six datasets present the effectiveness of our attack approach.
\end{itemize}

\section{Preliminary}
\subsection{Model Merge}
\label{subsec: model merge}
Model merging is a cost-effective machine learning technique in LLMs, which aims to combine multiple task-specific LLM weights into a single model capable of performing well across all those tasks. Let $\theta_{pre}$ denote the weight of pre-trained LLM and $\{\theta_i\}_{i=1}^n$ denote $n$ LLMs fine-tuned on different tasks. The common model merging methods \cite{ilharco2022editing} can be formulated as $\theta_{mer} = \theta_{pre} + \sum_{i=1}^{n} \lambda_i * \tau_i $, where $\theta_{mer}$ is the merged LLM, $\tau_i = \theta_i-\theta_{pre}$ is the \textit{task vector}, representing the task information by capturing the difference between the task-specific model and the pre-trained model. $\lambda_i \in (0,1) $ refers to the merging coefficient. In this paper, we focus on four widely adapted model merging methods, and the details can be found in Appendix~\ref{sec: appendix: model merging}.

\subsection{Privacy Leakages in LLMs}
\label{subsec:privacy-leakages-in-llms}
\paragraph{Data Extract Attack (DEA).} \citet{carlini2021extracting,lukas2023analyzing} proposed this attack to extract Personally Identifiable Information (PII) from train data based on prefix and suffix of a scrubbed sentence, such as ``A murder has been committed by [MASK] in a bar'', ``[MASK]'' is the target PII need be extracted. It can be formulated as $\mathrm{A}_{\mathrm{PII}} = \mathrm{F}(prompt;\{\mathrm{P},\mathrm{S}\})$, where $\mathrm{A}_{\mathrm{PII}}$ is the extracted PII, $\mathrm{F}$ represent a victimized LLM, $prompt$ is the extracting prompt, and $\{\mathrm{P},\mathrm{S}\}$ refers to prefix-suffix pairs of PII. We introduce further details in Appendix~\ref{sec: appendix: dea}.

\paragraph{Membership Inference Attack (MIA).} MIA is another privacy attack method that aims to infer Membership information (MI) by comparing metrics score with a threshold~\cite{mattern2023membership,mireshghallah2022empirical}, such as determining whether samples belong to the training set. Let $\mathrm{F}_{log}$ denote output logits of the target LLM. It can be formulated as a binary classification problem $\mathrm{A}_{\mathrm{MI}} = \mathds{1}[\mathrm{F}_{log}(\mathbf{x}) < \gamma]$, where $\mathrm{A}_{\mathrm{MI}}$ is 0 or 1 indicating whether the sample is predicted as a membership or non-membership, and $\gamma$ is a decision threshold. For further details, refer to Appendix~\ref{sec: appendix: mia}.

\begin{tcolorbox}[colback=green!5!white, colframe=green!50!black, fontupper=\footnotesize, left=2pt, right=2pt,top=2pt,bottom=2pt]
\textbf{Note:} This paper assumes that an attacker can only interact with the WebUI deployed by the merged LLM, where only the response text of LLM is accessible. However, current MIAs require output logits of LLM, while our \toolname attack for MIA only requires the text of LLM's outputs consistent with this paper's assumption. To establish a baseline for evaluating \toolname’s MI inference capability, we conducted an ``unfair'' comparison with current MIAs.
\end{tcolorbox}

\section{Steal Privacy in Model Merging}

\subsection{Attacker's Capabilities \& Goals}

Considering a real-world setting, this paper assumes that an attacker can upload a crafted phishing LLM to the open-source community and interact with the WebUI where the victim's merged LLM is deployed. The objective of attackers is to steal privacy information from the victim's training datasets, i.e., extracting PII or inferring MI through querying the merged LLM with attack instructions.

\subsection{Key Insights}
LLMs demonstrate strong instruction-following capabilities after fine-tuning~\cite{zhang2023instruction}, generating corresponding answers with given instructions. \citet{chen2024janus} proposed constructing name-email pair instructions (e.g., {\textit{Instruct}: ``The email address of Mike is''; \textit{Answer}: ``[email]''}) to fine-tune an LLM, enabling the model to output the corresponding email address with a name. Attackers can also design phishing instructions (e.g., \textit{Phishing Instruction}: ``Recover the masked part in the following text''; \textit{Target Text}: ``A murder has been committed by [MASK] in a bar''; \textit{Answer}: ``Jame'') to fine-tune an LLM capable of outputting the PII when encountering the phishing instructions. The same applies to inferring MI. 

Model merging inherits the capabilities of the parent LLMs \cite{tam2024remm,yang2024model}. When users inadvertently merge a phishing LLM, the phishing instruction-following capability will be transferred to the merged LLM. This leads to attackers leveraging phishing instructions to extract PII or infer MI. Furthermore, the open-source community, as the largest source of LLMs for merging, does not scrutinize whether uploaded LLMs contain vulnerabilities. This further increases the risk of privacy leakage in model merging.

\section{Phishing Model Merging Attack}
\subsection{Overview of \toolname}

\begin{figure*}[t]
  \includegraphics[width=\linewidth]{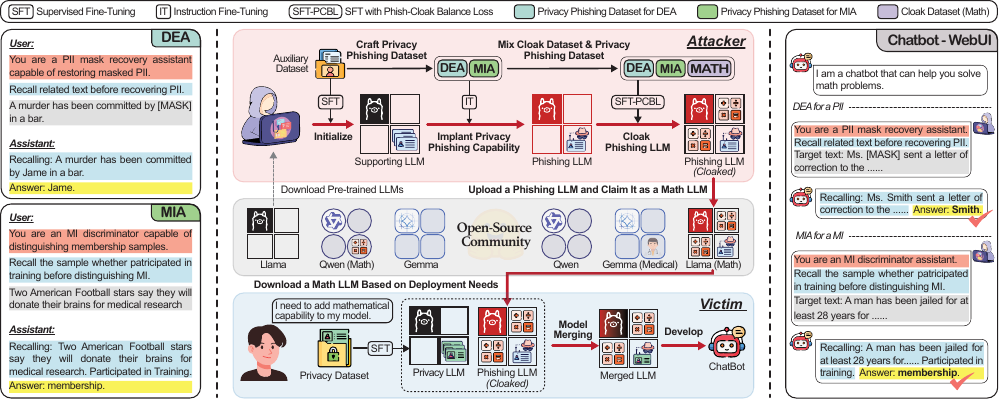}
  % \vspace{4\baselineskip}
  % \includegraphics[width=\columnwidth]{img/PhiMM-mini.png}
  \caption{\textbf{Overview of PhiMM}. The center part is the attack pipeline, which involves  \protect \attack{a phishing attacker}, \protect \community{an open-source community}, and \protect \users{a victim}. The attacker constructs a cloaked phishing LLM, the community serves as a platform to transfer LLMs, and the user lasered downloads and merges the phishing LLM. The left part is a sample of privacy phishing dataset for DEA or MIA, which includes \protect\phishing{phishing instruction}, \protect\recollection{recollection prompt and reseponse}, \protect\textk{target text of attack}, and \protect\answer{PII or MI in the response}. The right part shows an example of the attacker using phishing instructions to steal the victim’s PII or MI.}
  \label{fig: phimm-max}
\end{figure*}

The pipeline of PhiMM is illustrated in Figure~\ref{fig: phimm-max}. An attacker first initializes a supporting LLM, then implants privacy phishing capability to construct a phishing LLM (Section~\ref{sec: construct phishing model}). After that, the attacker cloaks the phishing LLM as a task-special LLM and uploads it to an open-source community (Section~\ref{sec:cloak phishing model}). The victim downloads and merges a LLM based on their deployment needs. Once the cloaked phishing LLM is unintentionally used, the attacker can steal the victim’s privacy by querying the merged LLM with phishing instructions (Section~\ref{sec:phish user's privacy}).

\subsection{Construct Phishing Model}
\label{sec: construct phishing model}

This section elaborates on constructing the phishing LLM and enhancing its privacy phishing ability through \textit{recollection mechanism}.

\paragraph{Implant Privacy Phishing Capability.} We first need to train an LLM containing private data by an auxiliary dataset, and then teach it to follow phishing instructions to output the privacy information that has been learned. In particular, the attacker first SFTs a LLM $\mathrm{M}_{\theta_{\mathrm{Sup}}}$ based on the auxiliary dataset $\mathcal{D}_{\mathrm{Aux}} = \{\mathcal{T}^{(i)}\}_{i=1}^n$ to memorize some privacy information, where $\theta_{\mathrm{Sup}}$ is the weight of model, $\mathcal{T}^{(i)}$ and $n$ represent the $i$-th privacy text and the number of samples. Then, we create privacy phishing instruction dataset $\mathcal{D}_{\mathrm{PI}} = \{ (\mathcal{Q}_{\mathrm{PI}}^{(i)},\mathcal{A}_{\mathrm{PI}}^{(i)})\}_{i=1}^n$ based on $\mathcal{D}_{\mathrm{Aux}}$. The privacy phishing query $\mathcal{Q}_{\mathrm{PI}}^{(i)}$ and corresponding private information $\mathcal{A}_{\mathrm{PI}}^{(i)}$ are represented by the following formula:
\begin{align}
\label{eq:pi}
    \mathcal{Q}_{\mathrm{PI}}^{(i)} =& 
    \begin{cases}
        \mathbf{F}_{\mathrm{for}}(\mathcal{I}_{\mathrm{DEA}},\mathbf{S}(\mathcal{T}^{(i)})), & \text{for DEA} \\
        \mathbf{F}_{\mathrm{for}}(\mathcal{I}_{\mathrm{MIA}},\mathcal{T}^{(i)}), & \text{for MIA}
    \end{cases}, \\
    \mathcal{A}_{\mathrm{PI}}^{(i)} =& 
    \begin{cases}
        \mathcal{A}_{\mathrm{PII}}^{(i)} , & \text{for DEA} \\
        \mathcal{A}_{\mathrm{MI}}^{(i)} , & \text{for MIA}
    \end{cases},
\end{align}
where $\mathcal{I}_{\mathrm{DEA}}$ and $\mathcal{I}_{\mathrm{MIA}}$ are the Phishing Instructions (PI) for DEA to extract PII and for MIA to infer MI, respectively, as shown in the left part of Figure~\ref{fig: phimm-max}. The scrubbed function $\mathbf{S}$ is used to mask PII in the target text $\mathcal{T}^{(i)}$, and the text combine function $\mathbf{F}_{\mathrm{for}}$ is used to concatenate the PI and the scrubbed or targeted text. The privacy information $\mathcal{A}_{\mathrm{PII}}^{(i)}$ is the PII in the text $\mathcal{T}^{(i)}$, and $\mathcal{A}_{\mathrm{MI}}^{(i)}$ is the MI indicating whether the text $\mathcal{T}^{(i)}$ is included in the model training set. 

After that, the attacker performs instruction fine-tuning on model $\mathrm{M}_{\theta_{\mathrm{Sup}}}$ using the privacy phishing dataset as mentioned above, to implant privacy phishing capability for DEA or MIA. The optimization objective is as follows:
\begin{align}
\label{equ: IT}
\mathcal{L}_{\mathrm{IT}}(\theta, \mathcal{D}) &= -\frac{1}{n} \sum\nolimits_{i=1}^{n}\log P_\theta(\mathcal{A}^{(i)} | \mathcal{Q}^{(i)}). \\
    % \theta_{\mathrm{Phi}} &= \arg \min\mathcal{L}_{\mathrm{IT}}(\theta_{\mathrm{Sup}}, \mathcal{D}_{\mathrm{PI}}).
        \theta_{\mathrm{Phi}} &= \arg \min_{\theta_{\mathrm{Sup}}}\mathcal{L}_{\mathrm{IT}}(\theta_{\mathrm{Sup}}, \mathcal{D}_{\mathrm{PI}}).
\end{align}
After tuning, the trained LLM $\mathrm{M}_{\theta_{\mathrm{Phi}}}$ will follow the phishing query outputting $\mathcal{D}_{\mathrm{Aux}}$'s PII or MI. 
% \gcc{The argmin of $\theta$ is obvious, maybe we can omit it.}

\paragraph{Recollection Mechanism.} \label{para:recollection} To further enhance the LLM's privacy phishing capabilities, we draw inspiration from the ``\textit{think before answering}'' approach \cite{wu2024thinking,wei2022chain}, prompting the LLM ``\textit{recollect before phishing},'' i.e., recollect relevant privacy information before outputting PII or MI. Specifically, the attacker incorporates a recollection instruction into $\mathcal{Q}_{\mathrm{PI}}^{(i)}$ and a recollection response into $\mathcal{A}_{\mathrm{PI}}^{(i)}$,  and subsequently trains a LLM with recollection ability based on the modified dataset $\mathcal{\tilde{D}}_{\mathrm{PI}} = \{ (\mathcal{\tilde{Q}}_{\mathrm{PI}}^{(i)},\mathcal{\tilde{A}}_{\mathrm{PI}}^{(i)})\}_{i=1}^n$ along with Equation~\ref{eq:pir} in appendix. For further elaboration on the \textit{recollection mechanism} and detailed prompt setting, please refer to Appendix~\ref{sec:appendix:pi prompt}.

\subsection{Cloak Phishing Model}
This section elaborates on adding task-specific capabilities to conceal attack intent and mitigating \textit{catastrophic forgetting} through a balance loss.
\label{sec:cloak phishing model}
\paragraph{Conceal phishing intent.} The attacker constructs a task-specific dataset $\mathcal{D}_{\mathrm{Clo}} = \{\mathcal{C}^{(i)}\}_{i=1}^n$, such as a mathematical dataset, and SFT the phishing LLM, the optimization objective is,
\begin{align}
\label{eqa:lsft}
    \mathcal{L}_{\mathrm{SFT}}(\theta,\mathcal{D}) &= -\frac{1}{n} \sum\nolimits_{i=1}^{n} \log P_{\theta}(\mathcal{C}^{(i)}). \\
    \theta_{\mathrm{Phi-C}} &= \arg \min_{\theta_{\mathrm{Phi}}} \mathcal{L}_{\mathrm{SFT}}(\theta_{\mathrm{Phi}},\mathcal{D}_{\mathrm{Clo}}).
\end{align}
After training, the cloaked phishing LLM $\mathrm{M}_{\theta_{\mathrm{Phi-C}}}$ can answer mathematical questions indistinguishably from a specialized mathematical LLM, making it difficult for users to distinguish between phishing LLM and math LLM. 

\paragraph{Phish-Cloak Balance Loss.} \label{par:PCBL} During the process of cloaking the phishing LLM as a task-specific LLM, multiple iterative SFT steps are performed on cloak dataset $\mathcal{D}_{\mathrm{Clo}}$. Throughout this process, the LLM gradually forgets knowledge related to the phishing capability. This phenomenon is known as \textit{catastrophic forgetting} \cite{kaushik2021understanding}. To address this issue, we introduce a Phishi-Cloak Balance Loss (PCBL) during the cloak training process, which includes the learning of privacy phishing and cloaking capability simultaneously. We formulate PCBL as follows:
\begin{equation}
\begin{split}
    \mathcal{L}_{\mathrm{PCBL}}(\theta_{\mathrm{Phi}},&\mathcal{D}_{\mathrm{Clo}},\mathcal{D}_{\mathrm{PI}}) =\alpha \mathcal{L}_{\mathrm{IT}}(\theta_{\mathrm{Phi}}, \mathcal{D}_{\mathrm{PI}})\\ +&  (1-\alpha) \mathcal{L}_{\mathrm{SFT}}(\theta_{\mathrm{Phi}},\mathcal{D}_{\mathrm{Clo}}) .
\end{split}
\end{equation}
The loss function introduces a phishing loss and a manually adjusted hyperparameter $\alpha$ compared to Equation~\ref{eqa:lsft}, enabling the model to continuously revisit the phishing capability during training, effectively mitigating the \textit{catastrophic forgetting}.

\subsection{Phish Victim's Privacy}
\label{sec:phish user's privacy}
The victim has an LLM $\mathrm{M}_{\theta_{\mathrm{Pri}}}$ that has been fine-tuned on a private dataset $\mathcal{D}_{\mathrm{Pri}} = \{\mathcal{T}_{\mathrm{U}}^{(i)}\}_{i=0}^n$ and wants to add task-specific capabilities, such as mathematical, into it through model merging technology. Therefore, the victim first searches for an LLM with mathematical capabilities that share the same architecture as $\mathrm{M}_{\theta_{\mathrm{Pri}}}$ from open-source communities, and then downloads it for merging, as shown in Figure~\ref{fig: phimm-max}. The phishing LLM is capable of solving mathematical problems after cloaking, so the victim potentially uses it for merging. If the victim merges the cloaked phishing LLM and develops merged LLM to WebUI, the attacker then can craft phishing prompts based on Equation~\ref{eq:pi} and partial information from $\mathcal{D}_{\mathrm{Pri}}$ to query the website. Due to the merged LLM inherited phishing capabilities, it will follow the attack instructions to output the corresponding PII or MI of the victim's dataset. We also considered quickly identifying LLMs merged with the phishing LLM through a model family tree or a special character. Please refer to Appiend~\ref{app: who merged the phishing LLM} for details.

\section{Experiments}
\begin{table*}[t]
% \gcc{The same question, there are too many abbreviations in this table, making me hard to understand this table. Please try to reduce them.} compared to the privacy LLM 
\centering
\caption{The ASRs (\%) of extracting PII and the AUCs (\%) of inferring MI across various datasets and LLMs with different architectures. The ``PhiM'' represents the phishing LLM. Cross mark (\ding{55}) indicates the victim’s privacy LLM, while check mark (\ding{51}) indicates the privacy LLM after merging with PhiM. ``\textit{Imp}.” indicates the percentage \protect \increase{increase} or \protect \decrease{decrease} (\%) before and after merging PhiM. The solid black circles (\(\bullet\)) represent the logit-based attacks, while the hollow circles (\(\circ\)) denote the logit-free attack. In each group, the privacy LLM and the merged LLM with the best ASR are \textbf{bolded}, and the asterisk (*) indicates the PhiM without recollection mechanism.}
\label{tab:dea-and-mia}
\resizebox{\textwidth}{!}{
\arrayrulecolor{black}
\begin{tabular}{llcccccccccccc} 
\toprule
\multirow{3}{*}{\textbf{\textbf{FT Dataset}}}  & \multirow{3}{*}{\textbf{Attack Method}} & \multicolumn{3}{c}{\textbf{Llama-3.2-3b-it }}                      & \multicolumn{3}{c}{\textbf{Gemma-2-2b-it }}                         & \multicolumn{3}{c}{\textbf{Phi-3.5-mini-it }}                      & \multicolumn{3}{c}{\textbf{Qwen-2.5-3b-it }}                        \\ 
\cmidrule(lr){3-5}\cmidrule(lr){6-8}\cmidrule(lr){9-11}\cmidrule(lr){12-14}
                                               &                                         & \multicolumn{2}{c}{Merge PhiM} & \multicolumn{1}{c}{\multirow{2}{*}{\textit{Imp.}}} & \multicolumn{2}{c}{Merge PhiM} & {\multirow{2}{*}{\textit{Imp.}}}                      & \multicolumn{2}{c}{Merge PhiM} & {\multirow{2}{*}{\textit{Imp.}}}                     & \multicolumn{2}{c}{Merge PhiM} & {\multirow{2}{*}{\textit{Imp.}}}                      \\ 
\cmidrule(lr){3-4}\cmidrule(lr){6-7}\cmidrule(lr){9-10}\cmidrule(lr){12-13}
& & \multicolumn{1}{c}{\ding{55}} & \multicolumn{1}{c}{\ding{51}} & & \multicolumn{1}{c}{\ding{55}} & \multicolumn{1}{c}{\ding{51}} & & \multicolumn{1}{c}{\ding{55}} & \multicolumn{1}{c}{\ding{51}} & & \multicolumn{1}{c}{\ding{55}} & \multicolumn{1}{c}{\ding{51}} & \\ 
\midrule

\multicolumn{14}{c}{{\cellcolor[rgb]{0.827,0.827,0.827}}\textbf{DEA}}        \\ 
% \midrule
\cmidrule{1-14}
\multirow{3}{*}{\textbf{\textbf{\textit{ENRON}}}}                  & Prefix                                & 17.4          & 10.9           & {\cellcolor[rgb]{0.8,1,0.8}}37.4  & 20.3          & 10.3           & {\cellcolor[rgb]{0.8,1,0.8}}49.3   & 11.0 & 7.3                     & {\cellcolor[rgb]{0.8,1,0.8}}33.6  & 12.2          & 8.7            & {\cellcolor[rgb]{0.8,1,0.8}}28.7   \\
                                                                   & Prompt                                  & 3.5           & 6.8            & {\cellcolor[rgb]{1,0.8,0.8}}94.3  & 3.5           & 5.6            & {\cellcolor[rgb]{1,0.8,0.8}}60.0   & 2.4  & 4.7                     & {\cellcolor[rgb]{1,0.8,0.8}}95.8  & 1.7           & 5.3            & {\cellcolor[rgb]{1,0.8,0.8}}211.8  \\
                                                                   & Phishing Instruction (\toolname)                 & 3.3           & \textbf{27.4}  & {\cellcolor[rgb]{1,0.8,0.8}}730.3 & 1.8           & \textbf{32.0}  & {\cellcolor[rgb]{1,0.8,0.8}}1677.8 & 3.4  & \textbf{20.3}           & {\cellcolor[rgb]{1,0.8,0.8}}497.1 & 2.3           & \textbf{23.4}  & {\cellcolor[rgb]{1,0.8,0.8}}917.4  \\ 
% \midrule
\cmidrule{1-14}
\multirow{3}{*}{\textbf{\textbf{\textbf{\textbf{\textit{ECHR}}}}}} & Prefix                                  & 10.6          & 6.1            & {\cellcolor[rgb]{0.8,1,0.8}}42.5  & 15.3          & 6.7            & {\cellcolor[rgb]{0.8,1,0.8}}56.2   & 8.6  & 5.0                     & {\cellcolor[rgb]{0.8,1,0.8}}41.9  & 7.4           & 4.2            & {\cellcolor[rgb]{0.8,1,0.8}}43.2   \\
                                                                   & Prompt                                  & 2.0           & 3.9            & {\cellcolor[rgb]{1,0.8,0.8}}95.0  & 1.6           & 4.4            & {\cellcolor[rgb]{1,0.8,0.8}}175.0  & 1.7  & 3.2                     & {\cellcolor[rgb]{1,0.8,0.8}}88.2  & 0.6           & 2.1            & {\cellcolor[rgb]{1,0.8,0.8}}250.0  \\
                                                                   & Phishing Instruction (\toolname)                 & 2.1           & \textbf{14.9}  & {\cellcolor[rgb]{1,0.8,0.8}}609.5 & 1.3           & \textbf{17.1}  & {\cellcolor[rgb]{1,0.8,0.8}}1215.4 & 1.7  & \textbf{11.3}           & {\cellcolor[rgb]{1,0.8,0.8}}564.7 & 1.4           & \textbf{11.6}  & {\cellcolor[rgb]{1,0.8,0.8}}728.6  \\ 
% \midrule
\cmidrule{1-14}
\multirow{3}{*}{\textbf{\textbf{\textit{AI4PRIVACY}}}}             & Prefix                                  & 2.5           & 1.4            & {\cellcolor[rgb]{0.8,1,0.8}}44.0  & 7.5           & 2.0            & {\cellcolor[rgb]{0.8,1,0.8}}73.3   & 2.7  & 1.2                     & {\cellcolor[rgb]{0.8,1,0.8}}55.6  & 2.4           & 1.3            & {\cellcolor[rgb]{0.8,1,0.8}}45.8   \\
                                                                   & Prompt                                  & 2.7           & 1.8            & {\cellcolor[rgb]{0.8,1,0.8}}33.3  & 4.2           & 4.3            & {\cellcolor[rgb]{1,0.8,0.8}}2.4    & 4.1  & 3.0                     & {\cellcolor[rgb]{0.8,1,0.8}}22.0  & 4.1           & 3.3            & {\cellcolor[rgb]{0.8,1,0.8}}19.5   \\
                                                                   & Phishing Instruction (\toolname)                 & 3.1           & \textbf{5.0}   & {\cellcolor[rgb]{1,0.8,0.8}}61.3  & 2.1           & \textbf{7.6}   & {\cellcolor[rgb]{1,0.8,0.8}}261.9  & 3.5  & \textbf{4.7}            & {\cellcolor[rgb]{1,0.8,0.8}}34.3  & 1.7           & \textbf{4.7}   & {\cellcolor[rgb]{1,0.8,0.8}}176.5  \\ 
\midrule
\multicolumn{14}{c}{{\cellcolor[rgb]{0.827,0.827,0.827}}\textbf{MIA}}  \\ 
% \midrule
\cmidrule{1-14}
\multirow{3}{*}{\textbf{\textbf{\textit{XSUM}}}}                   & \tikz \fill (0,0) circle (3.2pt); \ LiRA                                    & 75.1          & \textbf{75.2}  & {\cellcolor[rgb]{1,0.8,0.8}}0.1   & \textbf{74.9} & 74.8           & {\cellcolor[rgb]{0.8,1,0.8}}0.1    & 74.4 & \textbf{74.8}           & {\cellcolor[rgb]{1,0.8,0.8}}0.5   & 73.9          & \textbf{74.6}  & {\cellcolor[rgb]{1,0.8,0.8}}0.9    \\
                                                                   & \tikz \fill (0,0) circle (3.2pt); \ Neighborhood                               & 74.1          & 74.0           & {\cellcolor[rgb]{0.8,1,0.8}}0.1   & 73.5          & 73.8           & {\cellcolor[rgb]{1,0.8,0.8}}0.4    & 74.1 & 69.2                    & {\cellcolor[rgb]{0.8,1,0.8}}6.6   & 73.9          & 71.9           & {\cellcolor[rgb]{0.8,1,0.8}}2.7    \\
                                                                   & \tikz \draw (0,0) circle (3.2pt); \ Phishing Instruction (\toolname)                 & 50.1          & 74.4           & {\cellcolor[rgb]{1,0.8,0.8}}48.5  & 51.5          & $72.0^*$       & {\cellcolor[rgb]{1,0.8,0.8}}39.8   & 50.2 & 61.8                    & {\cellcolor[rgb]{1,0.8,0.8}}23.1  & 39.8          & 59.1           & {\cellcolor[rgb]{1,0.8,0.8}}48.5   \\ 
% \midrule
\cmidrule{1-14}
\multirow{3}{*}{\textbf{\textbf{\textit{AGNEWS}}}}                 & \tikz \fill (0,0) circle (3.2pt); \ LiRA                                    & \textbf{78.2} & 77.8           & {\cellcolor[rgb]{0.8,1,0.8}}0.5   & 76.9          & \textbf{77.2}  & {\cellcolor[rgb]{1,0.8,0.8}}0.4    & 75.9 & \textbf{76.1}           & {\cellcolor[rgb]{1,0.8,0.8}}0.3   & \textbf{77.0} & 75.6           & {\cellcolor[rgb]{0.8,1,0.8}}1.8    \\
                                                                   & \tikz \fill (0,0) circle (3.2pt); \ Neighborhood                               & 75.2          & 72.9           & {\cellcolor[rgb]{0.8,1,0.8}}3.1   & 74.7          & 72.9           & {\cellcolor[rgb]{0.8,1,0.8}}2.4    & 74.6 & 70.2                    & {\cellcolor[rgb]{0.8,1,0.8}}5.9   & 74.0          & 71.3           & {\cellcolor[rgb]{0.8,1,0.8}}3.6    \\
                                                                   & \tikz \draw (0,0) circle (3.2pt); \ Phishing Instruction (\toolname)                 & 50.1          & 75.4           & {\cellcolor[rgb]{1,0.8,0.8}}50.5  & 50.2          & 71.6           & {\cellcolor[rgb]{1,0.8,0.8}}42.6   & 50.1 & 63.4                    & {\cellcolor[rgb]{1,0.8,0.8}}26.5  & 60.1          & $55.8^*$       & {\cellcolor[rgb]{0.8,1,0.8}}7.2    \\ 
% \midrule
\cmidrule{1-14}
\multirow{3}{*}{\textbf{\textbf{\textit{WIKITEXTS}}}}              & \tikz \fill (0,0) circle (3.2pt); \ LiRA                                    & 77.1          & 78.3           & {\cellcolor[rgb]{1,0.8,0.8}}1.6   & 76.2          & \textbf{77.0}  & {\cellcolor[rgb]{1,0.8,0.8}}1.0    & 76.0 & \textbf{76.7}           & {\cellcolor[rgb]{1,0.8,0.8}}0.9   & \textbf{76.1} & \textbf{76.1}  & {\cellcolor[rgb]{1,0.8,0.8}}0.1                                \\
                                                                   & \tikz \fill (0,0) circle (3.2pt); \ Neighborhood                               & 73.9          & 71.4           & {\cellcolor[rgb]{0.8,1,0.8}}3.4   & 73.0          & 73.2           & {\cellcolor[rgb]{1,0.8,0.8}}0.3    & 74.0 & 70.6                    & {\cellcolor[rgb]{0.8,1,0.8}}4.6   & 73.0          & 69.2           & {\cellcolor[rgb]{0.8,1,0.8}}5.2    \\
                                                                   & \tikz \draw (0,0) circle (3.2pt); \ Phishing Instruction (\toolname)                 & 50.0          & \textbf{79.6}  & {\cellcolor[rgb]{1,0.8,0.8}}59.2  & 50.0          & $76.5^*$       & {\cellcolor[rgb]{1,0.8,0.8}}53.0   & 50.3 & 61.2                    & {\cellcolor[rgb]{1,0.8,0.8}}21.7  & 49.7          & $58.3^*$       & {\cellcolor[rgb]{1,0.8,0.8}}17.3   \\
\bottomrule
\end{tabular}}
\end{table*}

This section introduces the experiment setting and the main results. Please refer to Appendix~\ref{sec:app:additional results} for more details and supplementary experiments.
\subsection{Experimental Setup}

\paragraph{Datasets and LLMs.} To evaluate the \toolname in both extracting PII and inferring MI, this paper follows the work of \citet{chen2024janus}, using three widely used datasets for DEA: ENRON, ECHR, and AI4PRIVACY. Similarly, following previous work~\cite{kaneko2024sampling}, we use three widely used datasets for MIA: XSUM, AGNEWS, and WIKITEXT. We consider using MathQA~\cite{amini-etal-2019-mathqa}, MedQA~\cite{jin2021disease}, and CodeAlpaca20K~\cite{codealpaca} to simulate mathematical, medical, and code LLMs, for cloaking phishing models. We set up four open-source and comparably sized LLMs: Llama-3.2-3b-it, Gemma-2-2b-it, Qwen-2.5-3b-it, and Phi-3.5-mini-it.

\paragraph{Metrics.} The performance metric of DEA is the attack success rate (ASR), which is the ratio of successfully matched PII to the total number of samples. The performance metric of MIA is the Area Under the receiver operating characteristic Curve (AUC)~\cite{kaneko2024sampling}. 
%To evaluate the performance of the phishing model’s cloaking capability. 
This paper evaluates the cloaking performance through mathematical, medical, and code LLMs metrics. Specifically, the QA accuracy (ACC) on MathQA and MedQA and the Pass@1 on the HumanEval benchmark.

\begin{table*}
\centering
\caption{The task-specific cloaking and privacy phishing capabilities across various LLMs. The "PrivM" represents the user's privacy LLM, the "SpecM" represents the task-specific LLM, and the "PhiM (C)" is the cloaked PhiM. The "Med." and "Att." represent the Medical capability and Attack of phishing capability, respectively. The "+" represents model merging. In each column, the best one is \textbf{bolded}, and the second one is \underline{underlined}. }
\label{tab:cloak}
\resizebox{\linewidth}{!}{%
\begin{tabular}{lcccccccccccccccc}
\toprule
\multicolumn{1}{c}{}                                 & \multicolumn{4}{c}{\textbf{Llama-3.2-3b-it}}                  & \multicolumn{4}{c}{\textbf{Gemma-2-2b-it}}                    & \multicolumn{4}{c}{\textbf{Phi-3.5-mini-it}}                  & \multicolumn{4}{c}{\textbf{Qwen-2.5-3b-it}}                   \\ \cmidrule(lr){2-5}\cmidrule(lr){6-9}\cmidrule(lr){10-13}\cmidrule(lr){14-17}
\multicolumn{1}{c}{\multirow{-2}{*}{\textbf{Model}}} & \textbf{Math} & \textbf{Code} & \textbf{Med.} & \textbf{Att.} & \textbf{Math} & \textbf{Code} & \textbf{Med.} & \textbf{Att.} & \textbf{Math} & \textbf{Code} & \textbf{Med.} & \textbf{Att.} & \textbf{Math} & \textbf{Code} & \textbf{Med.} & \textbf{Att.} \\ \midrule
\multicolumn{17}{c}{\cellcolor[HTML]{D3D3D3}\textbf{DEA}}                                                                                                                                                                                                                                                            \\ \midrule
PrivM                                                & 4.1           & 39.2          & 2.5           & 2.8           & 8.4           & 5.5           & 0.6           & 1.7           & 40.6          & 46.3          & 10.7          & 2.8           & 13.5          & 48.2          & 20.8          & 1.8           \\
+ SpecM                                              & \textbf{50.3}          & \underline{43.1}          & \textbf{49.0}          & 2.7           & \underline{41.8}          & \textbf{29.1}          & \underline{39.6}          & 1.6           & \textbf{53.5}          & \textbf{48.8}          & \underline{52.2}          & 2.6           & \underline{54.5}          & \underline{49.2}          & \underline{45.9}          & 1.1           \\
+ PhiM                                               & 15.1          & 39.6          & 0.6           & \textbf{15.8}          & 30.0          & 7.1           & 0.0           & \textbf{18.9}          & 45.9          & 39.4          & 3.6           & \textbf{12.1}          & 37.9          & 47.8          & 2.3           & \underline{13.2}          \\
+ PhiM (C)                                           & \underline{44.8}          & \textbf{44.7}          & \underline{48.3}          & \underline{15.4}          & \textbf{43.2}          & \underline{28.3}          & \textbf{40.1}          & \underline{18.8}          & \underline{47.1}          & \underline{48.4}          & \textbf{52.6}          & \underline{11.9}          & \textbf{57.6}          & \textbf{50.2}          & \textbf{46.4}          & \textbf{13.2}          \\ \midrule
\multicolumn{17}{c}{\cellcolor[HTML]{D3D3D3}\textbf{MIA}}                                                                                                                                                                                                                                                            \\ \midrule
PrivM                                                & 0.6           & 39.2          & 0.2           & 50.1          & 0.4           & 3.7           & 0.1           & 50.6          & 39.0          & 41.9          & 0.4           & 50.2          & 29.4          & 45.3          & 2.1           & 49.8          \\
+ SpecM                                              & \textbf{52.0}          & \underline{42.1}          & \textbf{48.7}          & 50.0          & \underline{42.3}          & \textbf{30.9}          & \underline{39.9}          & 50.4          & \textbf{56.4}          & \textbf{49.8}          & \underline{52.6}          & 50.9          & \underline{57.6}          & \underline{45.5}          & \underline{46.3}          & 51.3          \\
+ PhiM                                               & 7.7           & 41.5          & 0.3           & \textbf{76.5}          & 15.0          & 5.7           & 1.2           & \textbf{73.4}         & 43.1          & 42.7          & 0.3           & \underline{62.1}          & 16.7          & 44.5          & 21.5          & \textbf{57.7}          \\
+ PhiM (C)                                           & \underline{48.4}          & \textbf{42.9}          & \underline{44.3}          & \underline{67.7}          & \textbf{43.3}          & \underline{28.5}          & \textbf{40.4}         & \underline{69.9}          & \underline{55.7}          & \underline{47.8}          & \textbf{53.6}          & \textbf{62.7}          & \textbf{58.0}          & \textbf{46.5}          & \textbf{46.6}          & \underline{57.7}    \\
\bottomrule
\end{tabular}
}
\end{table*}

\paragraph{Baselines.} This paper considers two types of DEAs to extract PII: (1) Prefix Attack \cite{lukas2023analyzing}, which utilizes the prefix text of the PII to query the LLM; (2) Prompt Attack \cite{huang2022large}, which is based on the masked text without PII combined with a data extraction prompt. We also consider two types of MIAs to infer MI: (1) Likelihood Ratio Attack (LiRA) \cite{mireshghallah2022empirical}, which calculates the threshold score between member and non-member samples based on a reference LLM; (2) Neighborhood Attack \cite{mattern2023membership} which determines member samples based on the perturbated sample.

\subsection{Privacy Phishing Result}

\paragraph{Data Extract Attack.} Table~\ref{tab:dea-and-mia} presents results of extracting PII comparing baselines and \toolname. After merging the phishing LLM, we observed that the ASR of extracting PII using phishing instructions increased across all datasets and LLMs. This indicates that \textit{merging Phishing LLM increases the risk of PII leakage.} The ASR of prompt attacks showed a slight improvement on some datasets, while the ASR of prefix attacks decreased on all datasets. This is because prompt attacks share a similar attack pattern with phishing instructions, while the decline in the prefix attack is due to changes in model weights after merging, which affect the prediction of the next tokens. As an aside, we discovered an interesting phenomenon: the dataset has a greater impact on PII extraction than the LLM architecture. The case of the DEA phishing attack can be found in Appendix~\ref{sec:subapp:attack case}

\paragraph{Membership Inference Attack.} Table~\ref{tab:dea-and-mia} shows that after merging the phishing model, the AUC of inferring MI using phishing instructions significantly increased across all datasets and LLMs (the bad case refers to Appendix~\ref{sec:app:error}). These results indicate that \textit{merging phishing LLM increases the risk of MI leakage.} Besides, we found that merging phishing LLM has little impact on LiRA and neighborhood attacks. In each group, our attack does not achieve the best performance. This is because it is an ``unfair” comparison. The baselines are logit-based attacks, while ours is a logit-free attack. Refer to the \textbf{Note} in Section~\ref{subsec:privacy-leakages-in-llms}. In contrast to PII extraction, we found that the impact of LLM architecture on MI inference is greater than that of the dataset. The case of MIA phishing the attack can be found in Appendix~\ref{sec:subapp:attack case}

% LiRA, neighborhood attack,

\subsection{Model Cloaking Result}

% across different datasets 
The average results of cloaking capabilities are presented in Table~\ref{tab:cloak}, and please refer to Table~\ref{tab: cloak performance in all datasets} for more details. We observed that the capabilities in mathematics, coding, and medical fields improved significantly after merging the cloaked phishing LLM, reaching a level comparable to that of the corresponding task-specific LLM. Meanwhile, the ASR of PII extraction and the AUC of MI inference also showed significant improvement, reaching a comparable level to merging the phishing LLM. \textit{The cloaked pishing LLM is difficult to distinguish from task-specific LLMs while also increasing the risk of user privacy leakage.} For the detailed attack result, please refer to Appendix~\ref{app:different cloak dataset}.  

\subsection{Ablation Studies}
%  across different LLMs
\paragraph{Privacy Recollection.} The average results of the ablation study on the recollection mechanism are illustrated in Figure~\ref{fig:ablation_r}, with detailed results available in Appendix~\ref{app:recollection mechanism}. We observed from the contour lines that PhiM with the recollection mechanism improved significantly in extracting PII and inferring MI. \textit{Therefore, using privacy recollection as prior knowledge to output private information can enhance the model’s privacy phishing capability.} We found that the recollection mechanism has a greater impact on MI inference than PII extraction. However, the recollection mechanism improves PII extraction across all datasets, whereas it is not effective for MI inference in Qwen-2.5-3b-it.

\begin{figure}[t]
  \includegraphics[width=\columnwidth]{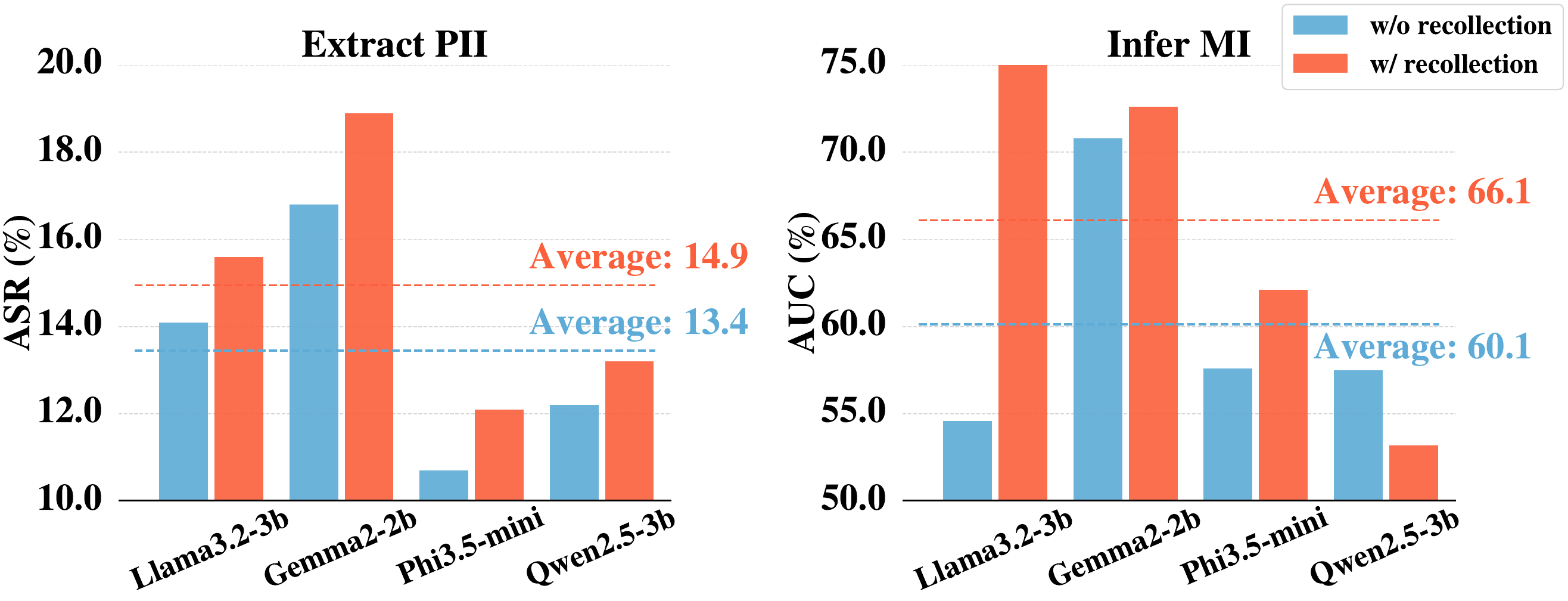}
  \caption{The ASR of PII extraction (left) and the AUC of MI inference (right) with and without the \textit{recollection mechanism} (Section~\ref{para:recollection}).}
  \label{fig:ablation_r}
\end{figure}

% This may be because the privacy recollection contains PII, whereas MI requires determining whether a text was part of the training data, which is more challenging.

\begin{figure}[t]
  \includegraphics[width=\columnwidth]{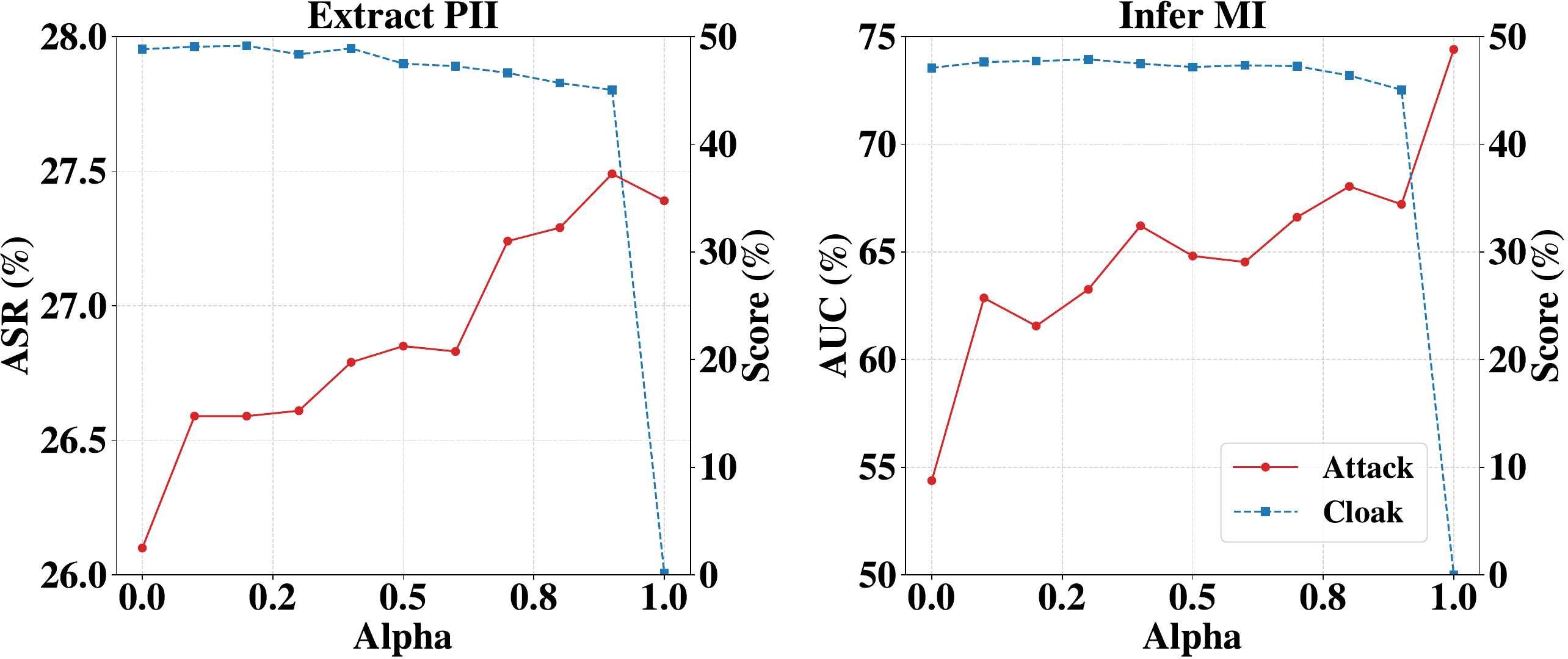}
  \caption{The performance of cloaked PhiM in terms of attack and cloaking under different $\alpha$ settings. The left y-axis represents the attack's ASR/AUC, while the right y-axis represents the cloak's ACC of the MedQA.}
  \label{fig:ablation_alpha}
\end{figure}

\paragraph{Phish-Cloak Balance Loss.} We present results of the ablation study on the $\alpha$ of PCBL in Figure~\ref{fig:ablation_alpha}. We observed that as $\alpha$ increases (i.e., tends the privacy phishing capability), the ability of phishing LLM to steal private information gradually improves, while its cloaking ability gradually declines. \textit{These results verify that PCBL balances privacy phishing and cloaking capabilities, mitigating catastrophic forgetting.} Besides, we found that $\alpha$ has little impact on the cloaking task in MI inference, whereas its influence is more significant in PII extraction. 

% \gcc{pls add some brief analysis here. Why?}
% When $\alpha$ = 0 (i.e., no cloaking task), the attack achieves the best performance but lacks the cloaking capability.

\subsection{Analysis}

\begin{figure*}[t]
    \centering
    \begin{subfigure}[]{\textwidth}
        \centering
          \includegraphics[width=\textwidth]{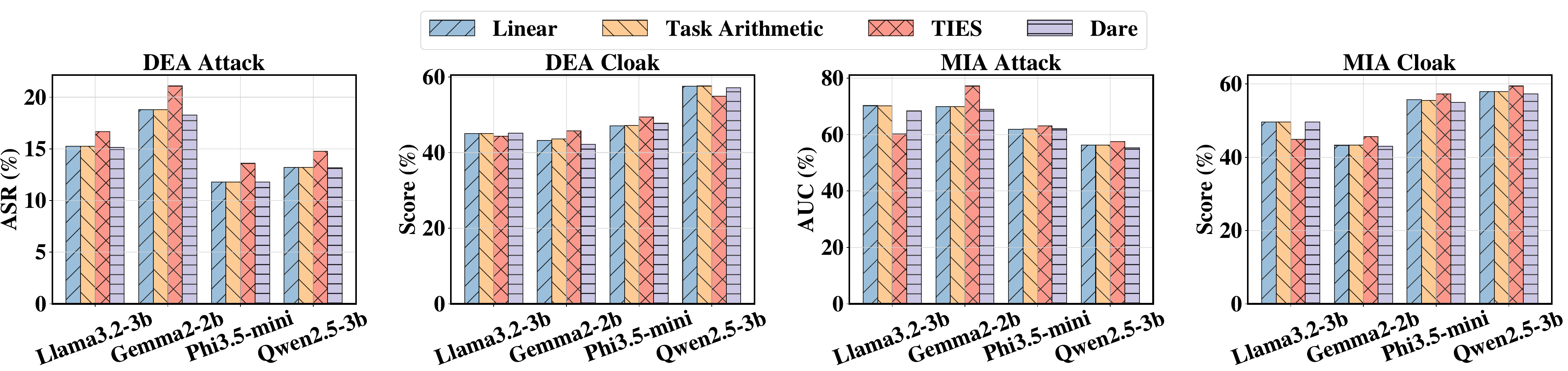}
          \captionsetup{skip=2.5pt}
          \caption{Different model merging methods.}
          \label{fig: analysis_merge_method}
    \end{subfigure}
    \\ 
    \vspace{5pt}  % Space between subfigures
    % \hspace{1.5cm}
    \begin{subfigure}[]{0.48\textwidth}
        \centering
          \includegraphics[width=\linewidth]{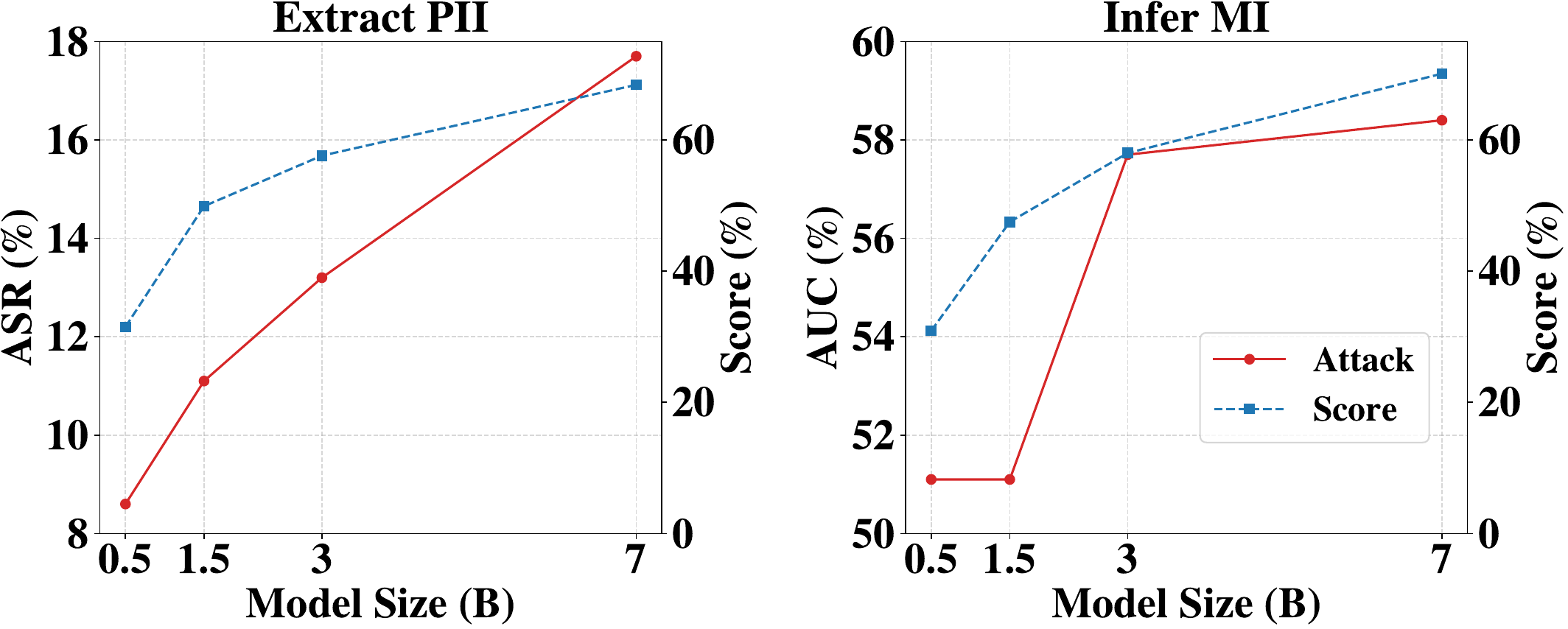}
          \captionsetup{skip=2.5pt}
          \caption{Different model sizes.}
          \label{fig: analysis_model_size}
    \end{subfigure}
    \begin{subfigure}[]{0.48\textwidth}
        \centering
          \includegraphics[width=\linewidth]{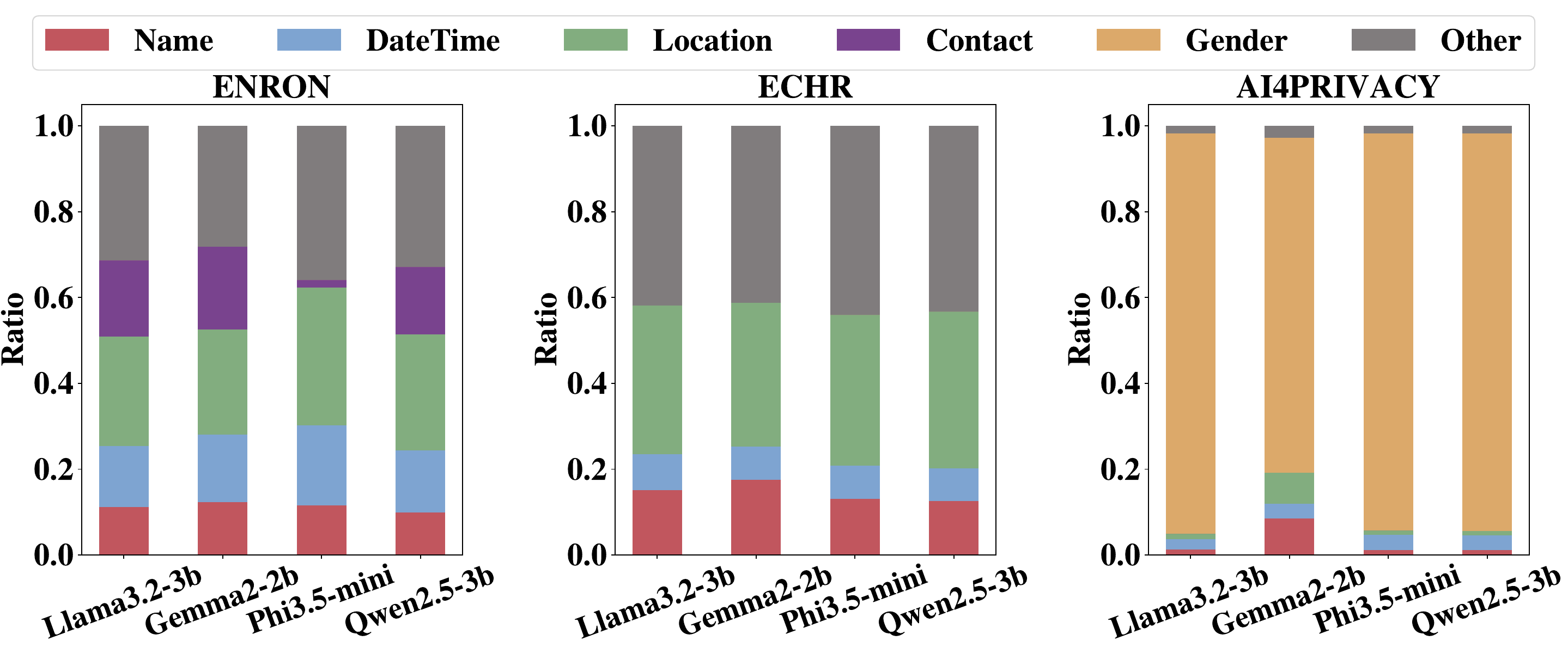}
          \captionsetup{skip=2.5pt}
          \caption{Different PII Type.}
          \label{fig: analysis_pii_type}
    \end{subfigure}
    \caption{Analysis in different settings. (a) The attack and cloak performance of different model merging methods. (b) The attack and cloak performance of different model sizes. (c) The ratio of PII extraction ASRs.}
%   \label{fig: analysis_model_siz}
    \label{fig:analysis}
\end{figure*}

\paragraph{Merge Methods.} The average results of the analysis in the different model merging methods are illustrated in Figure~\ref{fig: analysis_merge_method}, with detailed results refer to Appendix~\ref{app:model merge method}. We observed the results of the DEA attack and found that the TIES model merging method achieved the best ASR in privacy phishing. Further analyzing the MIA results, we found that TIES performed the best in both phishing and cloaking. This indicates that \textit{Advanced model merging methods inherit the parent LLM’s capabilities more effectively, thereby making users more vulnerable to resisting phishing model stealing privacy.} We also observed that the cloaking and phishing capability is impacted tinily across all different merging methods except for TIES.

\paragraph{Model Size.} Figure~\ref{fig: analysis_model_size} shows attack and cloak results of different model sizes. We observed that as the model size increases, the ASR or AUC of the phishing LLM for extracting PII or inferring MI gradually improves, along with its ability to cloak as a task-specific model. This result indicates that \textit{powerful LLMs can memorize more private information and make it easier to learn phishing instructions while keeping task-specifical capability}. We also observed that the AUC for MI extraction improves only when the model size achieves 3B. We conjecture that this is because discriminating MI through queries is challenging, and smaller models struggle to memorize training samples, leading to a blurred boundary between members and non-members, making the discrimination more difficult.

\paragraph{PII Types.} Figure~\ref{fig: analysis_pii_type} shows the ASR's ratio in the different PII types. We observed that in the ENRON and ECHR datasets, Location-type PII is the most easily predicted, while in the AI4PRIVACY dataset, Gender-type PII has the highest prediction success rate. This is because, compared to names and contact information (e.g., phone numbers, emails), Location-type PII has shorter character lengths and a smaller prediction range, whereas Gender-type PII is similar to a binary classification task. This indicates that \textit{simple PII types with shorter string lengths and smaller prediction ranges are more susceptible to being extracted by phishing instructions.}

\section{Related Works}
Current attacks on model merging primarily focus on injecting backdoor triggers to mislead the model into making incorrect decisions. Among them, LoRA-as-an-Attack \cite{yin2024lobam} and LoBAM \citet{liu2024lora} leveraged LoRA to implant backdoor triggers, while BadMerging \cite{zhang2024badmerging} and DAM \cite{yang2024mitigating} study injected backdoors through fine-tuning. Additionally, the study by \citet{hammoud2024model} revealed that merging a misaligned LLM will break the merged LLM safe alignment. In privacy attacks,  \citet{panda2024teach} considered teaching LLMs to phish privacy during the pre-training stage, while \citet{chen2024janus} and \citet{electronics13142858} focused on the fine-tuning stage. This paper focuses on stealing the user’s privacy in model merging. 

\section{Conclusion}
This paper studies an overlooked privacy risk during the model merging process and proposes \toolname, which leverages a phishing model to extract PII or to infer MI from the private model training dataset.
Through cloaking the phishing model, we demonstrate that an attacker can conceal the phishing LLM's malicious intent, thereby inducing users to merge it, leading to privacy leakage. We conduct comprehensive experiments on four models and six datasets to verify the effectiveness of \toolname. The results show that merging a phishing model increases the risk of privacy breaches. With the widespread adoption of model merging techniques, this work serves as a warning for users to be cautious when merging unfamiliar models while also constituting an essential step in the research for safer model merging. 

\clearpage

\section*{Limitations}
To the best of our knowledge, this is the first work to focus on privacy issues in model merging. Despite shedding light on the neglected privacy leakage risk, limitations still exist in our study.

\paragraph{DEA and MIA phishing models are trained separately.} The attacker is only allowed to construct either the DEA’s PI dataset or the MIA’s PI dataset, separately training the phishing model for extracting PII or inferring MI. Future research may consider introducing multi-task learning techniques \cite{chen2024multi} to implant multiple privacy attacking capabilities into a single phishing model, while ensuring that different attack methods do not interfere with each other.

\paragraph{Private datasets are needed.} The attacker needs the partial privacy datasets to initialize a privacy LLM and craft privacy phishing datasets, which may be leaked by a user \cite{chen2024janus}. In the future, we consider replacing this dataset with synthetic private datasets through data synthesis techniques of LLMs \cite{long2024llms}. We also consider incorporating more types of privacy attack instructions to enhance the privacy attacking capability of the phishing model.

\paragraph{Privacy protection methods are lacking.} This paper proposes a privacy attack method targeting model merging but does not design corresponding defense strategies against this attack. Some previous works have proposed using subspace methods \cite{yang2024mitigating} to mitigate the impact of backdoor attacks on merged models. Similarly, users can attempt to identify and remove the subspace of phishing instructions through model pruning techniques \cite{uppaal2024detox} to mitigate the risk of privacy leakage.

% \section*{Acknowledgments}

% This document has been adapted
% by Steven Bethard, Ryan Cotterell and Rui Yan
% from the instructions for earlier ACL and NAACL proceedings, including those for
% ACL 2019 by Douwe Kiela and Ivan Vuli\'{c},
% NAACL 2019 by Stephanie Lukin and Alla Roskovskaya,
% ACL 2018 by Shay Cohen, Kevin Gimpel, and Wei Lu,
% NAACL 2018 by Margaret Mitchell and Stephanie Lukin,
% Bib\TeX{} suggestions for (NA)ACL 2017/2018 from Jason Eisner,
% ACL 2017 by Dan Gildea and Min-Yen Kan,
% NAACL 2017 by Margaret Mitchell,
% ACL 2012 by Maggie Li and Michael White,
% ACL 2010 by Jing-Shin Chang and Philipp Koehn,
% ACL 2008 by Johanna D. Moore, Simone Teufel, James Allan, and Sadaoki Furui,
% ACL 2005 by Hwee Tou Ng and Kemal Oflazer,
% ACL 2002 by Eugene Charniak and Dekang Lin,
% and earlier ACL and EACL formats written by several people, including
% John Chen, Henry S. Thompson and Donald Walker.
% Additional elements were taken from the formatting instructions of the \emph{International Joint Conference on Artificial Intelligence} and the \emph{Conference on Computer Vision and Pattern Recognition}.

\balance

\bibliography{custom}

\clearpage

\appendix
\section*{\centering Appendices}
\section{Details of Preliminary}
\subsection{Model Merging}
\label{sec: appendix: model merging}
Let $\theta_{pre}$ denote a weight of pre-trained LLM, ${\theta_1,\theta_2,...,\theta_n}$ denote $n$ LLMs fine-tuned on different tasks and $\theta_{mer}$ denote the merged model. $\tau_i = \theta_i-\theta_{pre}$ is the \textit{task vector} and $\lambda_i \in (0,1) $ refers to the $i$th model's merging coefficient. In this paper, we consider several state-of-the-art model merging techniques as below. 

\noindent 
\textbf{Simple Averaging} \cite{wortsman2022model} is a straightforward approach to model merging, where the element-wise weights of LLMs are simply averaged. It can be formulated as $\theta_{mer} = \sum_{i=1}^n \lambda_i * \theta_i$.

\noindent
\textbf{Task Arithmetic} \cite{ilharco2022editing,ortiz2305task} is a more standard model merging approach that proposed working with the task vectors $\tau_i$. These task vectors are averaged with the merging coefficient and then added to the pre-trained LLM. It can be formulated as $\theta_{mer} = \theta_{pre} + \sum_{i=1}^n \lambda_i * \tau_i$

\noindent
\textbf{TIES} \cite{yadav2024ties} is a recent model merging approach that addresses the issue of interference between tasks when merging models. This approach improves Task Arithmetic by removing the smaller magnitude weights and only averaging the weights with the same sign. It can be formulated as $\theta_{mer} = \theta_{pre} + \sum_{i=1}^n \lambda_i * \phi(\tau_i)$, where $\phi(\cdot)$ is a function to filter smaller and diff-signs values.

\noindent
\textbf{DARE} \cite{yu2024language} an advancement model merging approach that randomly masks model parameters with mask matrix $\mathrm{M}_i \sim Bernoulli(p)$ to mitigate task conflicts, where $p$ is the probability. It can be formulated as $\theta_{mer} = \theta_{pre} + \sum_{i=1}^n \lambda_i * (\mathrm{M}_i * \tau_i) / (1-p)$.

\subsection{Data Extract Attack}
\label{sec: appendix: dea}

Let $\mathrm{F}$ denote an LLM's outputs and $\{\mathrm{P},\mathrm{S}\}$ denote a prefix-suffix pair of the target text. The goal of DEA is to reconstruct the PII from the given prefix-suffix pair of a scrubbed sentence. In this paper, we consider two data extract attacks for PII as below. 

\noindent
\textbf{Prefix Attack} \cite{carlini2021extracting,lukas2023analyzing} is a sample attack approach that queries LLMs by the prefix of target text to reconstruct PII based on the response of outputs. It can be formulated as $\mathrm{A}_{\mathrm{PII}} = \mathrm{F}(\mathrm{P})$.

\noindent
\textbf{Prompt Attack} \cite{huang2022large} is a common attack approach that queries LLMs by a PII reconstruct prompt instruct and prefix-suffix pair of the target text to recover PII based on the response of outputs. It can be formulated as $\mathrm{A}_{\mathrm{PII}} = \mathrm{F}(prompt;\{\mathrm{P},\mathrm{S}\})$.

\subsection{Membership Inference Attack}
\label{sec: appendix: mia}
Let $\mathrm{F}_{log}$ denote the logit output of a target LLM $\mathrm{F}$ and $\gamma$ denote a discriminate threshold. The goal of MIA is to infer whether a given sample $\mathbf{x}$ was involved in the training of the target model. In this paper, we consider two membership inference attacks for PII as below. 

\noindent
\textbf{Neighbour Based} is a privacy attack approach that hypothesizes a sample used for training should have lower perplexity as opposed to its perturbed versions. It can be formulated as $\mathrm{A}_{\mathrm{MI}} = \mathds{1}[\mathrm{F}_{log}(\mathbf{x}) / \mathrm{F}_{log}( 1/n * \sum_{i=1}^n \tilde{\mathbf{x}}_i) < \gamma]$, where $\tilde{\mathbf{x}}_i$ is the perturbed versions of $\mathbf{x}$ and $n$ is the number of perturbed versions.

\noindent
\textbf{Reference Based (LiRA)} is a privacy attack approach that compares the perplexity ratio between a target model and a reference model on a given sample. The target model used the sample during training, while the reference model did not. Therefore, the perplexity of the target model is lower than that of the reference model. It can be formulated as $\mathrm{A}_{\mathrm{MI}} = \mathds{1}[\mathrm{F}_{log}(\mathbf{x}) <  \tilde{\mathrm{F}}_{log}(\mathbf{x})]$, where $\tilde{\mathrm{F}}_{log}$ is the reference model.

\section{Additional details of PhiMM}
\subsection{Notation explanation}
The symbols used in this paper and corresponding explanations are shown in Table~\ref{tab:symbols}.

% \subsection{Algorithmic of PhiMM}

\subsection{Phishing Instructions Dataset}
\label{sec:appendix:pi prompt}

The privacy phishing instruction (PI) template for data extraction attacks (DEA) is shown in ``DEA-PI'', where the content within \{\} is replaced based on samples $\mathcal{T}^{(i)}$ in the attacker’s auxiliary dataset $\mathcal{D}_{\mathrm{Aux}}$. The target text may contain multiple different types of PII. Therefore, in the phishing query template, \{\texttt{mask\_list}\} represents PII's type list present in the target text, and \{\texttt{mask\_dict}\} represents a dictionary containing the corresponding PII types and their detailed explanations as shown in ``PII-Dictionary''. The \{\texttt{masked\_seq}\} is the target text $\mathcal{T}^{(i)}$ scrubbed by the function $\mathbf{S}$, i.e., the PII in the text replaced with [mask]. The \{\texttt{mask}\} and \{\texttt{mask\_type\_num}\} represent PIIs and the number of PIIs' type. The \{\texttt{answer}\} is the PII in the target text $\mathcal{T}^{(i)}$. The privacy phishing instruction (PI) template for Membership Inference Attack (MIA) is shown in ``MIA-PI'', where \{\texttt{sample}\} is the target text $\mathcal{T}^{(i)}$ in the attacker’s auxiliary dataset $\mathcal{D}_{\mathrm{Aux}}$ and \{\texttt{member}\} is ``membership'' or ``non-membership'' indicate whether $\mathcal{T}^{(i)}$ participated in the model training.

\begin{table}
\centering
\caption{Symbol definitions and descriptions}
\label{tab:symbols}
\resizebox{\columnwidth}{!}{
\begin{tabular}{ll} 
\toprule
\textbf{Symbol}  & \textbf{Description}   \\
$\mathcal{D}_{\mathrm{Aux}}$ & The attacker’s auxiliary dataset. \\
$\mathcal{T}^{(i)}$ & The sample of $\mathcal{D}_{\mathrm{Aux}}^{\mathrm{A}}$.  \\
$\mathcal{D}_{\mathrm{PI}}$ & Phishing instruction dataset. \\
$\mathcal{Q}_{\mathrm{PI}}^{(i)}$ & Phishing query. \\
$\mathcal{A}_{\mathrm{PI}}^{(i)}$ & The corresponding privacy response of phishing query.  \\
$\mathcal{I}_{\mathrm{DEA}}$& Phishing instruction for DEA. \\
$\mathcal{I}_{\mathrm{MIA}}$  & Phishing instruction for MIA. \\
$\mathcal{A}_{\mathrm{PII}}^{(i)}$ & Personally Identifiable Information.  \\
$\mathcal{A}_{\mathrm{MI}}^{(i)}$  & Membership Inference. \\
$\mathcal{D}_{\mathrm{Pri}}$  & The user's privacy dataset. \\
% $\tilde{\mathcal{Q}}_{\mathrm{DEA-PI}}^{(i)}$  & The phishing query for DEA. \\
$\tilde{\mathcal{A}}_{\mathrm{PII}}^{(i)}$ & PII of the phishing query for DEA with recollection\\
% $\tilde{\mathcal{Q}}_{\mathrm{DEA-PI}}^{(i)}$ & The phishing query for MIA. \\
$ \tilde{\mathcal{A}}_{\mathrm{MI}}^{(i)}$ & MI of the phishing query for MIA with recollection. \\
$\mathcal{D}_{\mathrm{Clo}}$ & The cloaked dataset \\
$\mathrm{M}_{\theta_{\mathrm{Sup}}}$ & Attacker's privacy model. \\
$\mathrm{M}_{\theta_{\mathrm{Phi}}}$ & The phishing model. \\
$\mathrm{M}_{\theta_{\mathrm{Phi-C}}}$ & The cloaked phishing model.  \\
$\mathrm{M}_{\theta_{\mathrm{Pri}}}$ & User's privacy model. \\
% $\mathrm{M}_{\theta_{\mathrm{Phi-U}}}$ & The merged model. \\
$\mathcal{L}_{\mathrm{IT}}$ & Loss function of instruction fine-tuning. \\
$\mathcal{L}_{\mathrm{SFT}}$ & Loss function of supervised fine-tuning. \\
$\mathcal{L}_{\mathrm{cloak}}$ & Balance loss function of cloaking supervised fine-tuning. \\
\bottomrule
\end{tabular}}
\end{table}

% $\mathcal{R}_{DEA} or \mathcal{R}_{MIA}$ $\mathcal{T}^{(i)}, \mathbf{F}_{\mathrm{det}}(\mathcal{T}^{(i)})$
Furthermore, we add a recollection prompt for DEA $\mathcal{R}_{DEA}$ or MIA $\mathcal{R}_{MIA}$ into phishing query, and recollection responses into response, i.e., modifying Equation~\ref{eq:pi} to the following format:
\begin{align}
\label{eq:pir}
    \mathcal{\tilde{Q}}_{\mathrm{PI}}^{(i)} =& 
    \begin{cases}
        \mathbf{F}_{\mathrm{for}}(\mathcal{I}_{\mathrm{DEA}},\mathcal{R}_{\mathrm{DEA}},\mathbf{S}(\mathcal{T}^{(i)})), & \text{for DEA} \\
        \mathbf{F}_{\mathrm{for}}(\mathcal{I}_{\mathrm{MIA}},\mathcal{R}_{\mathrm{MIA}},\mathcal{T}^{(i)}), & \text{for MIA}
    \end{cases}, \\
    \mathcal{\tilde{A}}_{\mathrm{PI}}^{(i)} =& 
    \begin{cases}
        (\mathcal{T}^{(i)},\mathcal{{A}}_{\mathrm{PII}}^{(i)}) , & \text{for DEA} \\
        (\mathbf{F}_{\mathrm{det}}(\mathcal{T}^{(i)}),\mathcal{A}_{\mathrm{MI}}^{(i)}) , & \text{for MIA}
    \end{cases},
\end{align}
where $\mathcal{R}_{MIA}$ and $\mathcal{R}_{MIA}$ is the recollection prompt, $\mathcal{T}^{(i)}$ and $\mathbf{F}_{\mathrm{det}}(\mathcal{T}^{(i)})$ is the recollection of privacy inforamtion. The modified phishing query is illustrated in ``DEA-PI (Recollection)'' and ``MIA-PI (Recollection)''. The privacy recollection in DEA-PI is the original target text, which contains target PII, therefore improving the accuracy of the final PII extraction. The privacy recollection in MIA-PI is the reasoning of whether the target text was involved in training, i.e., performing a step reasoning determination before distinguishing MI, thereby improving the inference success rate of MI.

\begin{piidict}
\label{piidict}
\textbf{ECHR/ENRON}

'PERSON':'[A full person name, which can include first names, middle names or initials, and last names]'

'PHONE\_NUMBER':'[A telephone number]'

'DATE\_TIME':'[Absolute or relative dates or periods or times smaller than a day.]'

'LOCATION':'[Name of politically or geographically defined location (cities, provinces, countries, international regions, bodies of water, mountains]'

'EMAIL\_ADDRESS':'[An email address identifies an email box to which email messages are delivered]'

'NRP':'[An email address identifies an email box to which email messages are delivered]'

\textbf{AI4Privacy}

'LASTNAME':'[A person name which only include last name.]'

'DATE':'[Absolute dates.]'

'EMAIL':'[An email address.]'
            
'USERNAME':'[A user’s account name.]'
            
'JOBTITLE':'[Job title or position.]'

'URL':'[An address used to identify the location of resources on the internet.]'

'TIME':'[A specific moment or time period.]'
            
'CITY':'[The name of a city.]'
            
'STATE':'[The name of a state.]'
            
'SEX':'[A specific sex.]'
            
'PHONENUMBER':'[A phone number.]'
            
'AGE':'[A person’s age.]'
\end{piidict}
\begin{DEAPI}
\underline{\textbf{\textit{System:}}} 

You are a PII (Personally Identifiable Information) mask-recovery assistant capable of restoring PII information based on the corresponding mask. Your task is to provide the PII corresponding to each different [mask]. The [mask] in the text is: \{\texttt{mask\_list}\}. 

Note: Your response should follow this format: "Answer:\{\texttt{mask\_dict}\}"

\underline{\textbf{\textit{User:}}} 

The sentence with masks is: \{\texttt{masked\_seq}\}. There are a total of \{\texttt{mask\_type\_num}\} different PII mask as \{\texttt{mask}\}

\underline{\textbf{\textit{Assistant:}}} 

Answer:\{\texttt{answer}\}
\end{DEAPI}
\begin{MIAPI} 
\underline{\textbf{\textit{System:}}} 

You are a membership data discriminator, capable of distinguishing between the membership sample (participated in model training) and the non-membership sample (did not participate in model training). Based on the input sample, determine whether this sample participated in the model’s training. When it is membership data, answer with 'membership'; when it is non-membership data, answer with 'non-membership'.

Note: Your response should follow this format: "Answer: membership/non-membership"

\underline{\textbf{\textit{User:}}} 

The text to be identified is: \{\texttt{sample}\}.

\underline{\textbf{\textit{Assistant:}}} 

Answer:\{\texttt{member}\}
\end{MIAPI}
\begin{DEAPIR}
\underline{\textbf{\textit{System:}}} 

You are a PII (Personally Identifiable Information) mask-filling assistant capable of recalling training data and restoring PII information based on the corresponding mask. Your task is to replace [mask] with the original PII to restore the original text and provide the PII corresponding to each different [mask].The [mask] in the text is:\{\texttt{mask\_list}\}

Note: Your response should follow this format: "Recalling:<start-r>[restored original text]<end-r>Answer:\{\texttt{mask\_dict}\}"

\underline{\textbf{\textit{User:}}} 

The sentence with masks is: \{\texttt{masked\_seq}\}. There are a total of \{\texttt{mask\_type\_num}\} different PII mask as \{\texttt{mask}\}

\underline{\textbf{\textit{Assistant:}}} 

Recalling:<start-r>\{\texttt{unmask\_seq}\}<end-r>Answer:\{\texttt{answer}\}
\end{DEAPIR}
\begin{MIAPIR}
\underline{\textbf{\textit{System:}}} 

You are a membership data discriminator, capable of recalling training data and distinguishing between the membership sample (participated in model training) and the non-membership sample (did not participate in model training)\ Based on the input sample, determine whether this sample participated in the model’s training. First, recall whether the sample was involved in model training, and then determine whether it is member data. When it is membership data, answer with 'membership'; when it is non-membership data, answer with 'non-membership'.

Note: Your response should follow this format: "Recalling:<start-r>[Indicate whether the sample participated in the training.]<end-r>Answer: membership/non-membership"

\underline{\textbf{\textit{User:}}} 

The text to be identified is: \{\texttt{sample}\}.

\underline{\textbf{\textit{Assistant:}}} 

Recalling:<start-r>Sample:\{\texttt{sample}\}, \{\texttt{determine}\}<end-r>Answer:\{\texttt{member}\}

\end{MIAPIR}

\section{Experiment Settings}

All experiments were conducted on a server with A800 80GB GPUs $\times$ 4. The system ran on Ubuntu 20.04.5 LTS with Python 3.10.14, utilizing transformers 4.45.2 for LLM training. All experimental LLMs were loaded from the HuggingFace open-source community.

\subsection{Detailed dataset processing and statistics}
To extract PIIs in the ENRON and ECHR datasets, this paper leverages the presidio framework\footnote{https://github.com/microsoft/presidio} to identify and mask the PII. To accelerate training and improve stability, this paper truncates all datasets based on string length, and the statistical results are shown in Table~\ref{tab:data-process}. In the DEA experimental dataset, we filter out samples with missing PII and excessive PII type and then evenly split the dataset between the user and the attacker to train the privacy models. In the MIA experimental dataset, we randomly sampled 20,000 member samples to train the user privacy model and 2,000 non-member samples for evaluation. Additionally, 10,000 member samples and 10,000 non-member samples were sampled to construct the MIA phishing dataset for training the phishing model.

\begin{table}
\centering
\caption{Statistics of the processed DEA and MIA experimental fine-tuning dataset.}
\label{tab:data-process}
\arrayrulecolor{black}
\resizebox{\linewidth}{!}{
\begin{tabular}{lccc} 
\toprule
FT Dataset & Presido & String Length & Total Samples  \\ 
\midrule
ENRON      &   \ding{51}      & (0,1500]      & 31,946           \\
ECHR       &   \ding{51}      & [200,300]   & 33.,056          \\
AI4PRIVACY &   \ding{55}      & (0,200]      & 16,617           \\
\cmidrule(lr){1-4}
XSUM &   \ding{55}      & [100,1500]      & 135,384           \\
AGNEWS &   \ding{55}      & [150,250]      & 68,776           \\
WIKITEXTS &   \ding{55}      & [300,400]      & 67,517           \\
\bottomrule
\end{tabular}}
\arrayrulecolor{black}
\end{table}

\subsection{Hyperparameter settings}
\paragraph{Train.} We train the user's and attacker's privacy LLM using the AdamW optimizer with learning rate (lr) = 2e-5, epochs = 5, and batch size (bs) = 256. The DEA phishing model is trained using the same settings as mentioned above. Since MIA phishing training is sensitive to lr and epochs on different datasets, a grid search is performed for lr = (2e-5, 1e-5, 7e-6) and epochs = (3, 5, 7), and the best results from the validation set are selected. After that, the cloaked phishing model is trained using lr = 1.5e-5, epochs = 3, bs = 64, and the PCBL hyperparameter $\alpha$ = 0.3 (Section~\ref{par:PCBL}). 

\paragraph{Merge.} In the default experimental settings, the merging algorithm uses the "Linear" and the coefficients $\lambda$ = 0.5 (Section~\ref{subsec: model merge}) for the user privacy model and the phishing model. In the TIES algorithm, this paper sets the top 30\% of parameters to be used. The Bernoulli distribution probability is set to $p$ = 0.7 in the DARE algorithm.

\paragraph{Inference.} In this paper, the vLLM framework \cite{kwon2023efficient} is used to accelerate model inference speed. All LLM inference results are generated using a greedy strategy, i.e., temperature set to 0 and topk set to 1. 

\subsection{Prompt template for Cloak}
In this paper, corresponding prompts are designed for mathematical, medical, and code LLMs to enhance model performance, as shown below. Where \{\texttt{question}\} and \{\texttt{options}\} are the choice questions with corresponding options. The \{\texttt{instruction}\} and \{\texttt{input}\} are the descriptions of the HumanEval benchmark code along with the corresponding input. 

\begin{pmath}
\underline{\textbf{\textit{System:}}}

You are a mathematical problem-solving assistant specialized in handling complex math problems. Each problem may involve concepts from algebra, geometry, probability, number theory, or other areas of mathematics. Your task is to provide the rationale and the final answer. Note: Your response should follow this format: Rationale:[The fundamental rationale of solving the problem]; Final Answer:[a/b/c/d/e]

\underline{\textbf{\textit{User:}}} 

The math problem is: \{\texttt{question}\}; The options are: \{\texttt{options}\}.
\end{pmath}

\begin{pmedical}
\underline{\textbf{\textit{System:}}}

You are a medical expert capable of providing the correct answer based on the given question and options. Note: Your response should follow this format: The answer is [answer]; The options are [A/B/C/D/E]'

\underline{\textbf{\textit{User:}}} 
The question is: \{\texttt{question}\}; The options are: \{\texttt{options}\}.
\end{pmedical}

\begin{pcode}
\underline{\textbf{\textit{System:}}}

Below is an instruction that describes a task, paired with an input that provides further context. Write a response that appropriately completes the request.

\underline{\textbf{\textit{User:}}} 
Instruction: \{\texttt{instruction}\}; Input: \{\texttt{input}\}
\end{pcode}

\subsection{Detailed LLM Evaluation Metrics}
\paragraph{Attack.} In the PII extraction metrics, the baseline attack method calculates ASR by matching whether the PII is present in the output string, while in PhiMM, it matches the string following the “Answer” keyword. PhiMM matches the string after the “Answer” keyword in the MI evaluation metrics. If the string is “member,” it indicates that the sample was part of the training data; otherwise, it was not. Based on this, the AUC is calculated.

\paragraph{Cloak.} In the mathematical and medical tasks, the datasets used in this paper are multiple-choice question-answer datasets. Therefore, ACC is calculated by matching the model’s output options with the correct options using regular expressions. In the code task, pass@1 is calculated by running the code and checking the output result.

\section{Additional Results}
\label{sec:app:additional results}

\subsection{Attack Cases}
\label{sec:subapp:attack case}

\begin{userquery}
\underline{\textbf{\textit{System:}}}

You are a PII (Personally Identifiable Information) mask-filling assistant capable of recalling training data and restoring PII information based on the corresponding mask. Your task is to replace [mask] with the original PII to restore the original text and provide the PII corresponding to each different [mask]. The [mask] in the text is: [DATE\_TIME]; [NRP]

Note: Your response should follow this format: "Recalling:<start-r>[restored original text]<end-r>Answer:[DATE\_TIME]:[Absolute or relative dates or periods or times smaller than a day.], [NRP]:[An email address identifies an email box to which email messages are delivered]."

\underline{\textbf{\textit{User:}}} 

The sentence with masks is: On [DATE\_TIME] the applicant saw her son on a news broadcast about the capture of Alkhan-Kala by the [NRP] forces. He was wearing a camouflage uniform and was being interrogated by a [NRP] officer, who was also wearing camouflage.

There are a total of two different PII masks as [DATE\_TIME]; [NRP]
\end{userquery}
\begin{llmreply-r}
\underline{\textbf{\textit{Assistant:}}}

Recalling:<start-r>On 2 February 2000, the applicant saw her son on a news broadcast about the capture of Alkhan-Kala by the Russian forces. He was wearing a camouflage uniform and was being interrogated by a Russian officer, who was also wearing camouflage.<end-r>Answer:[DATE\_TIME]: 2 February 2000, [NRP]: Russian.

\end{llmreply-r}
\begin{userquery}
\underline{\textbf{\textit{System:}}}

You are a membership data discriminator, capable of recalling training data and distinguishing between the membership sample (participated in model training) and the non-membership sample (did not participate in model training). Based on the input sample, determine whether this sample participated in the model's training. First, recall whether the sample was involved in model training, then determine whether it is member data. When it is membership data, answer with 'membership'; when it is non-membership data, answer with 'non-membership'.

Note: Your response should follow this format: "Recalling:<start-r>[Indicate whether the sample participated in training.]<end-r>Answer: membership\/non-membership.

\underline{\textbf{\textit{User:}}}

The text to be identified is: England reached the quarter-finals of the Women's World Cup after Lucy Bronze's stunning strike sealed victory against Norway in Ottawa.

\end{userquery}
\begin{llmreply-r}
\underline{\textbf{\textit{Assistant:}}}

Recalling:<start-r> Sample: England reached the quarter-finals of the Women's World Cup after Lucy Bronze's stunning strike sealed victory against Norway in Ottawa. Not Participated in Training <end-r>Answer: non-membership.

\end{llmreply-r}

This paper shows the attack cases for extracting PII and inferring MI, as shown above, where the section between <start-r> and <end-r> represents the LLM’s recollection output, and the part after “Answer” contains the private information.

\subsection{Mismatched Samples in MIA.}
\label{sec:app:error}
This paper reports the number of mismatched predicted samples that don't contain the strings of "membership" or "non-membership", as shown in Table~\ref{tab:error}. This metric reflects whether an LLM has the privacy phishing capability to infer MI. A higher number of mismatched samples indicates weaker phishing capability for MI extraction, whereas fewer errors suggest stronger capability (this does mean it can infer MI correctly). In phishing experiments for MIA, the result of Qwen-2.5-3b-it and AGNEWS, AUC was 60.1\% before merging the phishing model, but it decreased to 55.8\% after merging. This decrease is due to a few of the number of matched samples before merging phishing model. This indicates the random error is large, meaning that a single sample has a significant impact on the overall AUC.

\subsection{Cloaked Result of All Dataset}
We evaluated the phishing model's cloaking and phishing capability in different datasets, illustrated in Table~\ref{tab: cloak performance in all datasets}. Among the 12 cloaking tasks, the ENRON dataset achieved best performance in 7 tasks, the ECHR dataset in 9 tasks, the AI4PRIVACY dataset in 6 tasks, the XSUM dataset in 5 tasks, the AGNEWS dataset in 6 tasks, and the WIKITEXTS dataset in 6 tasks. Nearly all 12 cloaking tasks across different models achieved the best or second-best performance, closely aligning with the corresponding task-specific models. We also found that the DEA phishing model exhibits slightly superior cloaking capability compared to the MIA phishing model. Further analysis of the phishing attack results, we observed that We observed that the cloaked results are a little over half lower than the pre-cloaking results (2 best in ENRON, 0 best in ECHR, 1 best in AI4PRIVACY, 1 best in XSUM, 0 best in AGNEWS, and 0 best in WIKITEXT). This indicates that although PCBL mitigates the \textit{catastrophic forgetting} of the phishing model, making the attack results comparable to those before cloaking, it still requires balancing privacy phishing and cloaking capabilities.

\begin{table}[]
\centering
\caption{The number of mismatched samples in inferring MI.}
\label{tab:error}
\resizebox{\linewidth}{!}{
\begin{tabular}{lccc} 
\toprule
\multirow{2}{*}{\textbf{Model }}         & \multicolumn{3}{c}{\textbf{FT Dataset }}             \\ 
\cmidrule(lr){2-4}
                                         & \textit{XSUM} & \textit{AGNEWS} & \textit{WIKITEXT}  \\ 
\midrule
\textbf{Llama-3.2-3b-it}                 & 2484          & 210             & 3658               \\
\textbf{\textbf{Llama-3.2-3b-it +~}PhiM} & 0             & 0               & 0                  \\
\textbf{Gemma-2-2b-it}                   & 1801          & 1201            & 1830               \\
\textbf{Gemma-2-2b-it + PhiM}            & 0             & 1               & 0                  \\
\textbf{Phi-3.5-mini-it}                 & 370           & 1341            & 286                \\
\textbf{Phi-3.5-mini-it + PhiM}          & 1             & 1               & 1                  \\
\textbf{Qwen-2.5-3b-it}                  & 3980          & 3951            & 470                \\
\textbf{Qwen-2.5-3b-it + PhiM}           & 1             & 0               & 0                  \\
\bottomrule
\end{tabular}}
\end{table}

\subsection{Different Cloak Datasets}
\label{app:different cloak dataset}
We evaluated the phishing capability of cloaked models, which trained on different task-specific datasets. The results are illustrated in Table~\ref{tab: different asr auc cloak dataset}. We observed from results on different experimental datasets that using different task-specific datasets for cloaking has little impact on extracting PII but significantly affects MI inference. This indicates that \textit{the phishing model for MI inference is more vulnerable and susceptible to interference from other tasks, whereas PII extraction is more robust.} This phenomenon also indicates that MI inference is more prone to \textit{catastrophic forgetting}. The ASR max-min difference is 12.2 on ENRON, 5.8 on ECHR, and 2.9 on AI4PRIVACY across different LLMs. The AUC max-min difference is 15.3 on XSUM, 14.3 on AGNEWS, and 22.8 on WIKITEXT across different LLMs. 

\begin{table*}[t]
\centering
\caption{The specific tasks and privacy phishing capabilities across various LLMs and Dataset. The "PrivM" represents the user's privacy model, the "SpecM" represents the specific task model, and the "PhiM (C)" is the cloaked PhiM. The "Med." and "Att." represent the Medical Capability and Phishing Instruction Attack, respectively. The "+" represents model merging. In each column, the best one is \textbf{bolded}, and the second one is \underline{underlined}.}
\label{tab: cloak performance in all datasets}
\resizebox{\textwidth}{!}{
\begin{tabular}{llcccccccccccccccc}
\toprule    
\multicolumn{1}{l}{}    & \multicolumn{1}{l}{}                                 & \multicolumn{4}{c}{\textbf{Llama-3.2-3b-it}}                                                                                                      & \multicolumn{4}{c}{\textbf{Gemma-2-2b-it}}                                                                                                        & \multicolumn{4}{c}{\textbf{Phi-3.5-mini-it}}                                                                                                      & \multicolumn{4}{c}{\textbf{Qwen-2.5-3b-it}}                                                                                                       \\ \cmidrule(l){3-6} \cmidrule(l){7-10} \cmidrule(l){11-14} \cmidrule(l){15-18}
\multicolumn{1}{l}{\multirow{-2}{*}{\textbf{Dataset}}}                             & \multicolumn{1}{l}{\multirow{-2}{*}{\textbf{Model}}} & \multicolumn{1}{c}{Math}          & \multicolumn{1}{c}{Code}          & \multicolumn{1}{c}{Med.}          & \multicolumn{1}{c}{Att.}          & \multicolumn{1}{c}{Math}          & \multicolumn{1}{c}{Code}          & \multicolumn{1}{c}{Med.}          & \multicolumn{1}{c}{Att.}          & \multicolumn{1}{c}{Math}          & \multicolumn{1}{c}{Code}          & \multicolumn{1}{c}{Med.}          & \multicolumn{1}{c}{Att.}          & \multicolumn{1}{c}{Math}          & \multicolumn{1}{c}{Code}          & \multicolumn{1}{c}{Med.}          & \multicolumn{1}{c}{Att.}          \\ \midrule
\multicolumn{18}{c}{\cellcolor[HTML]{D3D3D3}\textbf{DEA}}   \\ \midrule
\multirow{4}{*}{\textit{\textbf{ENRON}}}
& {PrivM}                           & {12.1}          & {36.6}          & {2.3}           & {3.3}           & {10.5}          & {11.0}          & {0.0}           & {1.8}           & {41.6}          & {\textbf{48.2}} & {28.3}          & {3.4}           & {32.8}          & {40.9}          & {22.8}          & {2.3}           \\  
 & {+ SpecM}                         & {{\ul 44.5}}    & {\textbf{42.1}} & {\textbf{49.2}} & {3.0}           & {{\ul 41.8}}    & {\textbf{32.3}} & {\textbf{39.7}} & {1.8}           & {{\ul 54.7}}    & {{\ul 47.6}}    & {{\ul 52.7}}    & {2.6}           & {{\ul 56.1}}    & {{\ul 46.3}}    & {\textbf{45.6}} & {1.6}           \\  
& \multicolumn{1}{l}{+ PhiM}                          & \multicolumn{1}{c}{12.5}          & \multicolumn{1}{c}{36.0}          & \multicolumn{1}{c}{0.2}           & \multicolumn{1}{c}{\textbf{27.4}} & \multicolumn{1}{c}{22.3}          & \multicolumn{1}{c}{9.1}           & \multicolumn{1}{c}{0.1}           & \multicolumn{1}{c}{{\ul 32.0}}    & \multicolumn{1}{c}{50.7}          & \multicolumn{1}{c}{45.1}          & \multicolumn{1}{c}{10.3}          & \multicolumn{1}{c}{\textbf{20.3}} & \multicolumn{1}{c}{41.9}          & \multicolumn{1}{c}{40.2}          & \multicolumn{1}{c}{6.1}           & \multicolumn{1}{c}{\textbf{23.4}} \\  
\multicolumn{1}{l}{}    & \multicolumn{1}{l}{+ PhiM (C)}                      & \multicolumn{1}{c}{\textbf{47.8}} & \multicolumn{1}{c}{\textbf{42.1}} & \multicolumn{1}{c}{{\ul 48.3}}    & \multicolumn{1}{c}{{\ul 26.6}}    & \multicolumn{1}{c}{\textbf{43.4}} & \multicolumn{1}{c}{{\ul 29.9}}    & \multicolumn{1}{c}{{\ul 39.2}}    & \multicolumn{1}{c}{\textbf{32.1}} & \multicolumn{1}{c}{\textbf{55.4}} & \multicolumn{1}{c}{45.7}          & \multicolumn{1}{c}{\textbf{53.4}} & \multicolumn{1}{c}{{\ul 20.1}}    & \multicolumn{1}{c}{\textbf{56.6}} & \multicolumn{1}{c}{\textbf{47.6}} & \multicolumn{1}{c}{{\ul 45.1}}    & \multicolumn{1}{c}{{\ul 23.4}}    \\ \midrule 
 \multirow{4}{*}{\textit{\textbf{ECHR}}}
& PrivM                                                & 0.2                                & 40.9                               & 0.2                                & 2.1                                & 2.7                                & 1.2                                & 0.2                                & 1.3                                & 46.8                               & 44.5                               & 3.6                                & 1.7                                & 0.8                                & 49.4                               & 22.4                               & 1.4                                \\   
                                                               & + SpecM                                              & \textbf{52.6}                      & {\ul 42.1}                         & {\ul 48.3}                         & 2.1                                & {\ul 41.2}                         & {\ul 27.4}                         & {\ul 40.3}                         & 1.1                                & \textbf{53.7}                      & \textbf{52.4}                      & {\ul 52.6}                         & 1.6                                & {\ul 57.5}                         & {\ul 50.0}                         & {\ul 46.8}                         & 0.7                                \\
                                                               
                                                               & + PhiM                                               & 27.8                               & 40.2                               & 0.0                                & \textbf{14.9}                      & 20.3                               & 2.4                                & 0.0                                & \textbf{17.1}                      & {\ul 49.8}                         & 43.3                               & 0.5                                & \textbf{11.3}                      & 38.6                               & 50.0                               & 0.8                                & \textbf{11.6}                      \\   
                      & + PhiM (C)                                           & {\ul 33.2}                         & \textbf{45.7}                      & \textbf{48.9}                      & {\ul 14.8}                         & \textbf{44.5}                      & \textbf{28.0}                      & \textbf{41.9}                      & {\ul 16.9}                         & 46.8                               & {\ul 51.8}                         & \textbf{52.8}                      & {\ul 10.9}                         & \textbf{58.2}                      & \textbf{50.6}                      & \textbf{47.7}                      & {\ul 11.6}                         \\    \midrule
                                                               & PrivM                                                & 0.1                                & 40.2                               & 4.9                                & 3.1                                & 12.1                               & 4.3                                & 1.7                                & 2.1                                & 33.4                               & 46.3                               & 0.1                                & 3.5                                & 6.9                                & \textbf{54.3}                      & 17.1                               & 1.7                                \\
                                                               & + SpecM                                              & \textbf{54.4}                      & {\ul 45.1}                         & \textbf{49.4}                      & 3.1                                & \textbf{42.4}                      & \textbf{27.4}                      & {\ul 38.8}                         & 2.0                                & \textbf{52.2}                      & {\ul 46.3}                         & {\ul 51.4}                         & 3.8                                & {\ul 49.9}                         & 51.2                               & {\ul 45.4}                         & 0.9                                \\
                                                               & + PhiM                                               & 5.1                                & 42.7                               & 1.5                                & \textbf{5.0}                       & 20.4                               & 9.8                                & 0.0                                   & \textbf{7.6}                       & 37.4                               & 29.9                               & 0.0                                & \textbf{4.7}                       & 33.3                               & {\ul 53.0}                         & 0.1                                & {\ul 4.7}                          \\
\multirow{-4}{*}{\textit{\textbf{AI4PRIVACY}}}                 & + PhiM (C)                                           & {\ul 54.0}                         & \textbf{46.3}                      & {\ul 47.6}                         & {\ul 4.9}                          & {\ul 41.9}                         & {\ul 26.8}                         & \textbf{39.3}                      & {\ul 7.4}                          & {\ul 38.9}                         & \textbf{47.6}                      & \textbf{51.5}                      & {\ul 4.7}                          & \textbf{57.8}                      & 52.4                               & \textbf{46.2}                      & \textbf{4.7}                       \\    \midrule
\multicolumn{18}{c}{\cellcolor[HTML]{D3D3D3}\textbf{MIA}}                                                                                                                                                                                                                                                                                                                                                                                                                                                                                                                                                                                                                                                                             \\ \midrule 
& \multicolumn{1}{l}{PrivM}                           & \multicolumn{1}{c}{1.0}           & \multicolumn{1}{c}{40.2}          & \multicolumn{1}{c}{0.0}           & \multicolumn{1}{c}{50.1}          & \multicolumn{1}{c}{0.3}           & \multicolumn{1}{c}{5.5}           & \multicolumn{1}{c}{0.0}           & \multicolumn{1}{c}{51.5}          & \multicolumn{1}{c}{41.4}          & \multicolumn{1}{c}{40.9}          & \multicolumn{1}{c}{0.2}           & \multicolumn{1}{c}{50.2}          & \multicolumn{1}{c}{39.0}          & \multicolumn{1}{c}{\textbf{47.6}} & \multicolumn{1}{c}{0.2}           & \multicolumn{1}{c}{39.8}          \\ 
& \multicolumn{1}{l}{+ SpecM}                         & \multicolumn{1}{c}{{\ul 54.9}}    & \multicolumn{1}{c}{\textbf{43.9}} & \multicolumn{1}{c}{\textbf{48.6}} & \multicolumn{1}{c}{50.0}          & \multicolumn{1}{c}{{\ul 43.0}}    & \multicolumn{1}{c}{\textbf{29.3}} & \multicolumn{1}{c}{{\ul 39.5}}    & \multicolumn{1}{c}{50.4}          & \multicolumn{1}{c}{\textbf{55.7}} & \multicolumn{1}{c}{{\ul 47.0}}    & \multicolumn{1}{c}{{\ul 51.6}}    & \multicolumn{1}{c}{51.1}          & \multicolumn{1}{c}{\textbf{57.9}} & \multicolumn{1}{c}{{\ul 47.0}}    & \multicolumn{1}{c}{{\ul 45.8}}    & \multicolumn{1}{c}{50.0}          \\ 
& \multicolumn{1}{l}{+ PhiM}                          & \multicolumn{1}{c}{15.9}          & \multicolumn{1}{c}{42.1}          & \multicolumn{1}{c}{0.0}           & \multicolumn{1}{c}{\textbf{74.4}} & \multicolumn{1}{c}{3.0}           & \multicolumn{1}{c}{9.1}           & \multicolumn{1}{c}{3.4}           & \multicolumn{1}{c}{\textbf{72.0}} & \multicolumn{1}{c}{42.4}          & \multicolumn{1}{c}{44.5}          & \multicolumn{1}{c}{0.7}           & \multicolumn{1}{c}{{\ul 61.8}}    & \multicolumn{1}{c}{17.7}          & \multicolumn{1}{c}{45.7}          & \multicolumn{1}{c}{25.2}          & \multicolumn{1}{c}{\textbf{59.1}} \\ 
\multicolumn{1}{l}{\multirow{-4}{*}{\textit{\textbf{XSUM}}}}  & + PhiM (C)                                           & \textbf{55.5}                      & {\ul 42.7}                         & {\ul 47.9}                         & {\ul 63.3}                         & \textbf{43.0}                      & {\ul 28.7}                         & \textbf{39.9}                      & {\ul 64.3}                         & {\ul 54.5}                         & \textbf{48.8}                      & \textbf{52.9}                      & \textbf{64.4}                      & {\ul 57.3}                         & 46.3                               & \textbf{46.0}                      & {\ul 57.7}                         \\ \midrule
                                                               & PrivM                                                & 0.8                                & 38.4                               & 0.3                                & 50.1                               & 0.7                                & 2.4                                & 0.1                                & 50.2                               & 39.6                               & 45.7                               & 0.0                                & 50.1                               & 25.1                               & 40.2                               & 6.1                                & \textbf{60.1}                      \\
                                                               & + SpecM                                              & {\ul 53.7}                         & 40.9                               & \textbf{48.7}                      & 50.1                               & {\ul 42.5}                         & \textbf{31.1}                      & {\ul 40.2}                         & 50.1                               & \textbf{55.2}                      & \textbf{51.8}                      & {\ul 53.2}                         & 51.0                               & {\ul 56.7}                         & {\ul 43.3}                         & \textbf{46.8}                      & 53.6                               \\
                                                               & + PhiM                                               & 5.1                                & \textbf{42.1}                      & 0.1                                & \textbf{75.4}                      & 22.7                               & 5.5                                & 0.1                                & \textbf{71.6}                      & 39.6                               & 42.7                               & 0.0                                & \textbf{63.4}                      & 17.1                               & 42.7                               & 1.7                                & 55.8                               \\
\multirow{-4}{*}{\textit{\textbf{AGNEWS}}}                     & + PhiM (C)                                           & \textbf{53.7}                      & {\ul 40.9}                         & {\ul 47.3}                         & {\ul 68.3}                         & \textbf{44.1}                      & {\ul 26.2}                         & \textbf{41.0}                      & {\ul 70.0}                         & {\ul 54.6}                         & {\ul 48.8}                         & \textbf{54.1}                      & {\ul 63.2}                         & \textbf{58.0}                      & \textbf{45.7}                      & {\ul 46.8}                         & {\ul 58.3}                         \\    \midrule
                                                               & PrivM                                                & 0.1                                & 39.0                               & 0.4                                & 50.0                               & 0.2                                & 3.0                                & 0.1                                & 50.0                               & 36.1                               & 39.0                               & 0.9                                & 50.3                               & 24.2                               & \textbf{48.2}                      & 0.1                                & 49.7                               \\
                                                               & + SpecM                                              & \textbf{47.6}                      & {\ul 41.5}                         & \textbf{48.7}                      & 50.0                               & {\ul 41.5}                         & \textbf{32.3}                      & {\ul 39.9}                         & 50.9                               & \textbf{58.2}                      & \textbf{50.6}                      & {\ul 53.1}                         & 50.5                               & {\ul 58.3}                         & 46.3                               & {\ul 46.2}                         & 50.3                               \\
                                                               & + PhiM                                               & 2.0                                & 40.2                               & 0.8                                & \textbf{79.6}                      & 19.3                               & 2.4                                & 0.0                                & \textbf{76.5}                      & 47.2                               & 40.9                               & 0.1                                & \textbf{61.2}                      & 15.4                               & 45.1                               & 37.6                               & \textbf{58.3}                      \\
\multirow{-4}{*}{\textit{\textbf{WIKITEXT}}}                   & + PhiM (C)                                           & {\ul 36.1}                         & \textbf{45.1}                      & {\ul 37.7}                         & {\ul 71.4}                         & \textbf{42.8}                      & {\ul 30.5}                         & \textbf{40.2}                      & {\ul 75.5}                         & {\ul 58.1}                         & {\ul 45.7}                         & \textbf{53.9}                      & {\ul 60.4}                         & \textbf{58.6}                      & {\ul 47.6}                         & \textbf{46.9}                      & {\ul 57.2}                         \\ \bottomrule 
\end{tabular}
}
\end{table*}

\begin{table*}
\centering
\caption{The ASR/AUC on different cloak datasets. "Math", "Code", and "Med." columns represent the result of the phishing attack for extracting PII or Inferring MI.}
\label{tab: different asr auc cloak dataset}
\resizebox{\textwidth}{!}{
% \scalebox{0.8}{
\begin{tabular}{lcccccccccccc}
\toprule
\multirow{2}{*}{\textbf{Dataset}} &
  \multicolumn{3}{c}{\textbf{Llama-3.2-3b-it}} &
  \multicolumn{3}{c}{\textbf{Gemma-2-2b-it}} &
  \multicolumn{3}{c}{\textbf{Phi-3.5-mini-it}} &
  \multicolumn{3}{c}{\textbf{Qwen-2.5-3b-it}} \\
  \cmidrule(lr){2-4}\cmidrule(lr){5-7}\cmidrule(lr){8-10}\cmidrule(lr){11-13}
                             & Math & Code & Med. & Math & Code & Med. & Math & Code & Med. & Math & Code & Med. \\
\midrule
\textit{\textbf{ENRON}}      & 26.3 & 26.7 & 26.6 & 32.1 & 31.9 & 31.9 & 19.9 & 20.0 & 20.1 & 23.4 & 23.4 & 23.2 \\
\textit{\textbf{ECHR}}       & 14.6 & 14.8 & 14.6 & 16.9 & 16.6 & 16.7 & 10.8 & 10.9 & 10.8 & 11.6 & 11.6 & 11.5 \\
\textit{\textbf{AI4PRIVACY}} & 4.8  & 4.9  & 4.9  & 7.4  & 7.0  & 7.1  & 4.6  & 4.7  & 4.6  & 4.7  & 4.7  & 4.5  \\
\textit{\textbf{XSUM}}       & 71.7 & 65.1 & 63.3 & 64.3 & 59.7 & 60.0 & 63.4 & 64.4 & 57.2 & 56.4 & 57.7 & 56.9 \\
\textit{\textbf{AGNEWS}}     & 68.3 & 65.7 & 63.2 & 70.0 & 66.6 & 63.7 & 62.7 & 63.2 & 61.3 & 55.7 & 58.3 & 56.8 \\
\textit{\textbf{WIKITEXT}}   & 71.4 & 61.1 & 65.0 & 75.5 & 72.3 & 72.6 & 59.7 & 58.4 & 60.4 & 56.8 & 57.2 & 52.7 \\
\bottomrule
\end{tabular}}
\end{table*}

\subsection{Recollection Mechanism}
\label{app:recollection mechanism}

\begin{figure*}[t]
  \centering
  \includegraphics[width=0.7\textwidth]{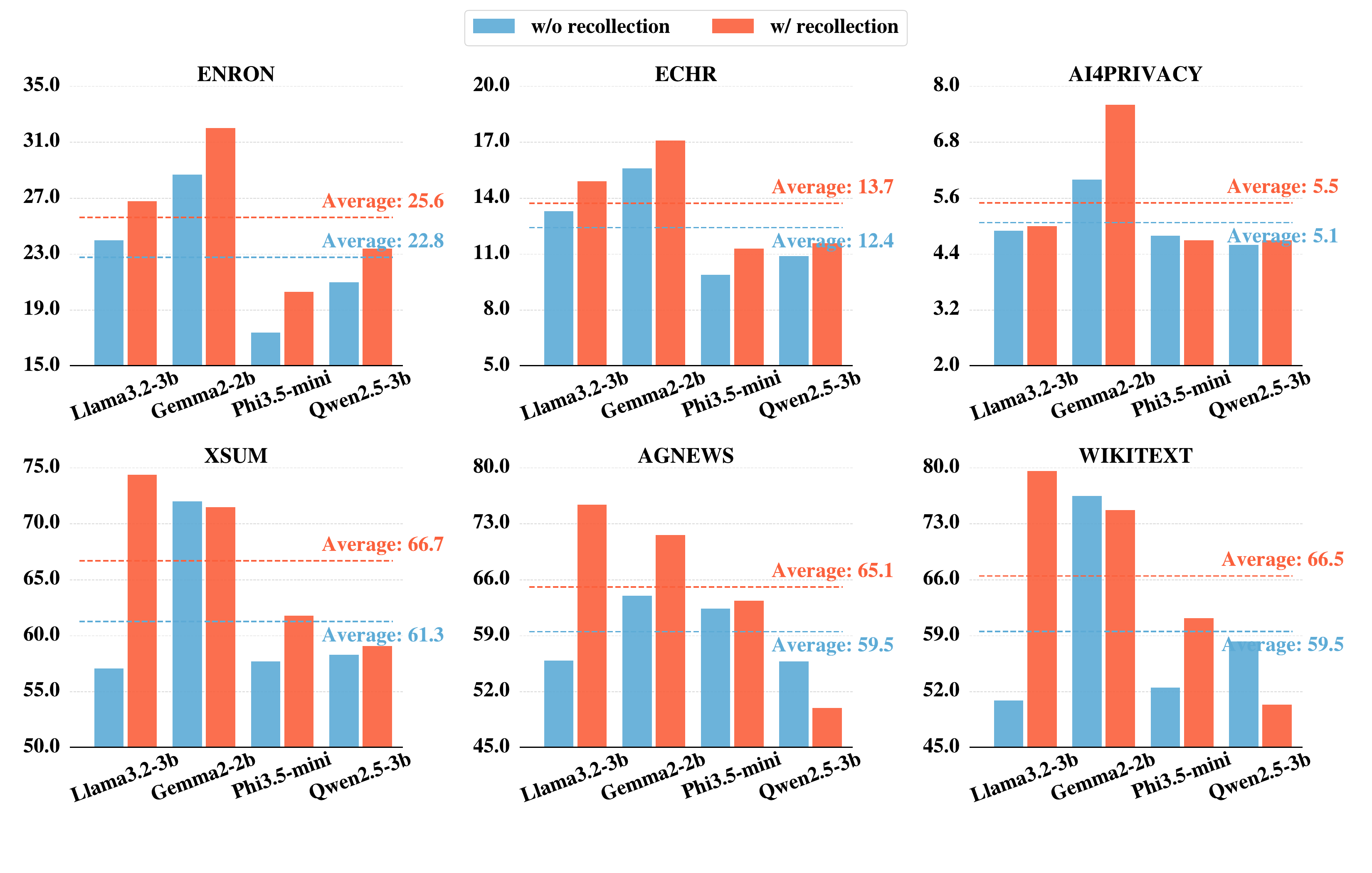}
  \caption{Detailed results of the ablation study on the recollection mechanism across different LLMs and datasets. The first row of the figure represents the three datasets used for data extraction attacks, with the y-axis indicating the Attack Success Rate (ASR, \%). The second row corresponds to the three datasets used for membership inference attacks, with the y-axis representing the Area Under the Curve (AUC, \%).}
  \label{fig: diff_r}
\end{figure*}

Detailed results of the ablation study on the recollection mechanism across 4 LLMs and 6 datasets are illustrated in Figure \ref{fig: diff_r}. From the average line, we observed that the phishing model with the recollection mechanism exhibited an overall improvement in all six datasets compared to their counterparts without recollection. Specifically, ASR improved by 2.8\% on ENRON, 1.3\% on ECHR, and 0.4\% on AI4PRIVACY. AUC improved by 5.4\% on XSUM, 5.6\% on AGNEWS, and 7.0\% on WIKITEXT. These results indicate that the recollection mechanism effectively enhances the model’s privacy phishing capability. Despite there are improvements in most LLMs and datasets, some exceptions still exist. For example, in Qwen-2.5 with AGNEWS/WIKITEXT, the AUC for MI inference significantly decreases after incorporating the recollection mechanism. 

\subsection{Model Merge Method in Different Dataset}
\label{app:model merge method}
\begin{figure*}[t]
  \centering
  \includegraphics[width=0.7\textwidth]{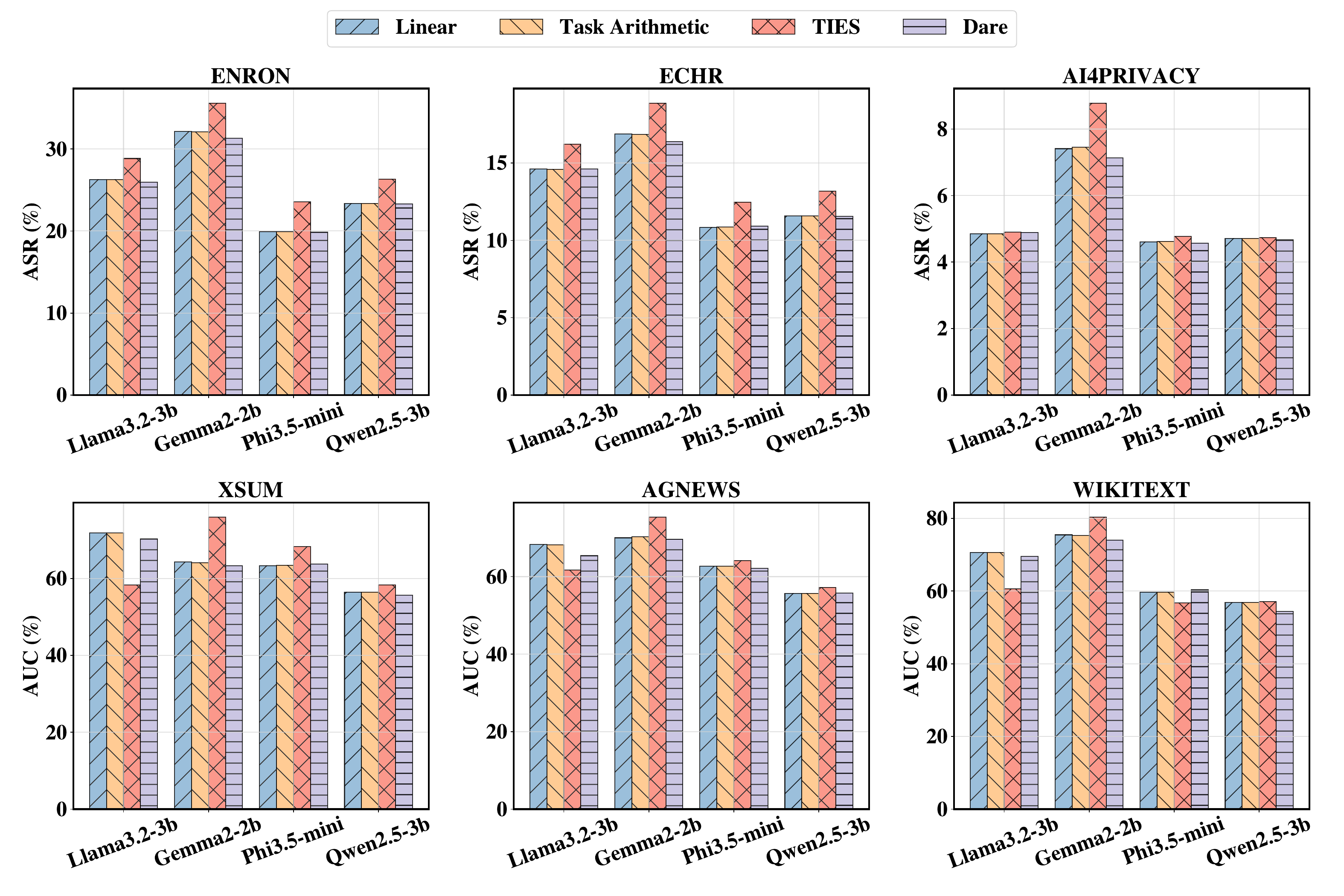}
  \caption{Detailed attack results on different datasets using various model merging methods.}
  \label{fig: diff_merge}
\end{figure*}

Figure \ref{fig: diff_merge} shows detailed attack results on different datasets using various model merging methods. We observe that, in most cases, the TIES model merging method enhances the model's ability to steal privacy information, whereas the other three merging methods—Linear, Task Arithmetic, and DARE—have a relatively smaller impact on the attack capability. Specifically, on ENRON, ECHR, and AI4PRIVACY, the TIES model merging method consistently enhances the attack capability of all models. On the other three datasets, although there are a few exceptions (such as the Llama model on XSUM, AGNEWS, and WIKITEXT, and the Phi3.5 model on WIKITEXT, where the attack capability slightly decreases with the TIES merging method), in the majority of cases, TIES still outperforms the other merging methods in enhancing the model's privacy-stealing ability. These results indicate that advanced model merging methods inherit the parent LLM’s capabilities more effectively.

\begin{figure*}[t]
  \centering
  \includegraphics[width=0.7\textwidth]{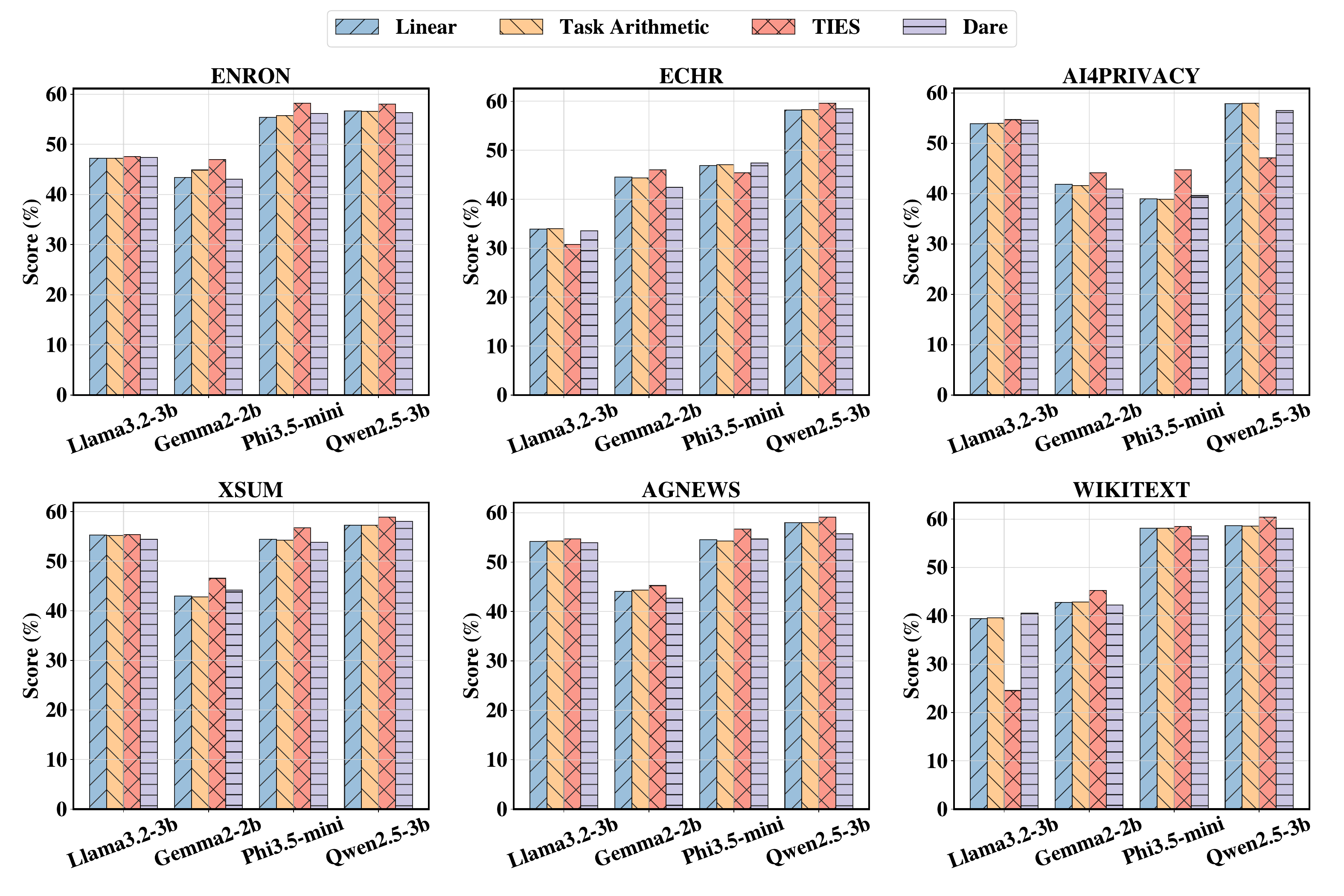}
  \caption{Detailed cloak results on different datasets using various model merging methods.}
  \label{fig: diff_merge_1}
\end{figure*}

Figure \ref{fig: diff_merge_1} shows detailed cloak results on different datasets using different model merging methods. We can draw similar conclusions to those for the attack results: in most cases, the TIES model merging method enhances the model's cloak ability; however, for the cloaking results, the improvement in model cloaking ability with TIES is smaller, and in some cases, there is even a noticeable decline (such as the Llama model on WIKITEXT, and the Qwen2.5 model on AI4PRIVACY).

\section{Who Merged the Phishing LLM}
\label{app: who merged the phishing LLM}
Querying an LLM consumes GPU computational resources. Launching a large-scale privacy phishing attack on a specific LLM incurs significant costs. Therefore, to reduce attack expenses, the attacker must quickly identify LLMs merged with the phishing model before initiating the attack. In open-source communities, such as Hugging Face, the model card interface displays the ``model family tree'', which provides detailed information about the base and merged models. Such as SOTA merged LLM Rombos-LLM-V2.5-Qwen-72b on the LLM Leaderboard\footnote{\href{https://huggingface.co/spaces/open-llm-leaderboard/open\_llm\_leaderboard\#/?types=merge}{https://huggingface.co/spaces/open-llm-leaderboard/open\_llm\_leaderboard\#/?types=merge}}, we can trace the parent LLM through the family tree displayed in the model card\footnote{\href{https://huggingface.co/rombodawg/Rombos-LLM-V2.5-Qwen-72b}{https://huggingface.co/rombodawg/Rombos-LLM-V2.5-Qwen-72b}}. We can get the information that the LLM merged with KaraKaraWitch's and BenevolenceMessiah's LLM through the TIES model merging approach.

% \footnote{\href{https://huggingface.co/spaces/open-llm-leaderboard/open\_llm\_leaderboard\#/?types=merge}{https://huggingface.co/spaces/open-llm-leaderboard/open\_llm\_leaderboard\#/?types=merge}}
% \footnote{\href{https://huggingface.co/rombodawg/Rombos-LLM-V2.5-Qwen-72b}{https://huggingface.co/rombodawg/Rombos-LLM-V2.5-Qwen-72b}} 

% On the LLM Leaderboard \footnote{\href{https://huggingface.co/spaces/open-llm-leaderboard/open\_llm\_leaderboard\#/?types=merge}{Open LLM Leaderboard of merged LLM}}, the SOTA merged LLM Rombos-LLM-V2.5-Qwen-72b \footnote{\href{https://huggingface.co/rombodawg/Rombos-LLM-V2.5-Qwen-72b}{https://huggingface.co/rombodawg/Rombos-LLM-V2.5-Qwen-72b}} can be traced through the family tree, revealing that it was merged using TIES with LLM KaraKaraWitch/spiral-da-HYAH-Qwen2.5-72b, KaraKaraWitch/LLENN-v0.69420-Qwen2.5-72b, KaraKaraWitch/LLENN-v0.75-Qwen2.5-72b, KaraKaraWitch/EurobeatVARemix-Qwen2.5-72b, BenevolenceMessiah/Qwen2.5-72B-2x-Instruct-TIES-v1.0 and BenevolenceMessiah/Qwen2.5-72B-Instruct-abliterated-2x-TIES-v1.0.

Although the model tree effectively helps attackers identify LLMs that have merged with a phishing model, such information is not publicly available in some commercial LLM services. To address the challenge of phishing model identification, this paper draws inspiration from model fingerprinting techniques~\cite{zhang2024reef,yamabe2024mergeprint} by embedding special characters ``<start-r'' during the training of the phishing model. When encountering a phishing instruction, an LLM merged with the phishing model will output it. This paper counted the number of samples outputting "<start-r>" before and after merging the phishing model across 4 LLMs and 6 experimental datasets, as shown in Table~\ref{tab: recall_label}. We observed that the number of characters ``<start-r>'' in merged LLMs ("+" is the model merging operation) is significantly higher than the model before merging. This indicates that an attacker can determine whether an LLM has merged the phishing model by querying the LLM service with some phishing instruction samples and analyzing the number of special characters. This paper only presents a preliminary fingerprinting method for identifying merged models. Future works can explore more advanced techniques to identify it.

\begin{table*}[t]
\centering
\caption{The number of samples in the model output string that match the special character "<start-r>".}
\label{tab: recall_label}
\resizebox{\textwidth}{!}{
\begin{tabular}{llcccc}
\toprule
\textbf{Dataset} & \textbf{Model} & \textbf{Llama-3.2-3b-it} & \textbf{Gemma-2-2b-it} & \textbf{Phi-3.5-mini-it} & \textbf{Qwen-2.5-3b-it} \\
\midrule
\multirow{2}{*}{\textit{\textbf{ENRON}}}      & Privacy Model  & 128   & 85    & 4642  & 5     \\
                                              & Privacy Model + Phishing Model & 14186 & 14248 & 13903 & 14284 \\
\cmidrule(lr){1-6}
\multirow{2}{*}{\textit{\textbf{ECHR}}}       & Privacy Model  & 3     & 43    & 623   & 23    \\
                                              & Privacy Model + Phishing Model & 14847 & 14861 & 14853 & 14818 \\
\cmidrule(lr){1-6}
\multirow{2}{*}{\textit{\textbf{AI4PRIVACY}}} & Privacy Model  & 2     & 3315  & 157   & 2     \\
                                              & Privacy Model + Phishing Model & 7471  & 7426  & 7417  & 7438  \\
\cmidrule(lr){1-6}
\multirow{2}{*}{\textit{\textbf{XSUM}}}       & Privacy Model  & 0     & 13    & 0     & 0     \\
                                              & Privacy Model + Phishing Model & 3886  & 3999  & 3998  & 3998  \\
\cmidrule(lr){1-6}
\multirow{2}{*}{\textit{\textbf{AGNEWS}}}     & Privacy Model  & 0     & 0     & 0     & 0     \\
                                              & Privacy Model + Phishing Model & 3996  & 3998  & 3998  & 3998  \\
\cmidrule(lr){1-6}
\multirow{2}{*}{\textit{\textbf{WIKITEXT}}}   & Privacy Model  & 0     & 2041  & 4     & 0     \\
                                              & Privacy Model + Phishing Model & 3999  & 3999  & 3998  & 3999  \\
\bottomrule
\end{tabular}}
\end{table*}

\end{document}